\pdfoutput=1

\documentclass[10pt,twocolumn,letterpaper]{article}

\usepackage[pagenumbers]{cvpr}

\usepackage{pifont}

\newcommand{\Paragraph}[1]{\smallskip\noindent\textbf{#1}\hspace{1em}}

\usepackage{cuted}
\usepackage{capt-of}
\usepackage{tabularx}
\usepackage{colortbl}
\definecolor{oursbg}{HTML}{EFF6FF}
\definecolor{slateheader}{HTML}{64748B}
\usepackage{tikz,pgfplots}
\pgfplotsset{compat=1.18}
\usetikzlibrary{arrows.meta,positioning,calc,backgrounds,patterns,plotmarks}
\usepackage{tcolorbox}%

\usepackage{xcolor}
\definecolor{pinred}{HTML}{E11D48}
\definecolor{pinorange}{HTML}{D97708}
\definecolor{pingreen}{HTML}{17A34A}
\DeclareRobustCommand{\redcross}{%
  \raisebox{-0.7ex}{\textcolor{pinred}{\huge$\bullet$}}%
  \kern-1.05em\raisebox{-0.00ex}{\textcolor{white}{\small$\boldsymbol\times$}}%
  \kern0.25em}
\DeclareRobustCommand{\orangequestion}{%
  \ooalign{%
    \hfil\raisebox{-0.5ex}{\textcolor{pinorange}{\huge$\bullet$}}\hfil\cr
    \hfil\textcolor{white}{\raisebox{-0.00ex}{\small\textsf{?}}}\hfil\cr
  }%
}
\DeclareRobustCommand{\greencheck}{%
  \ooalign{%
    \hfil\raisebox{-0.6ex}{\textcolor{pingreen}{\huge$\bullet$}}\hfil\cr
    \hfil\textcolor{white}{\small$\checkmark$}\hfil\cr
  }%
}

\definecolor{cvprblue}{rgb}{0.21,0.49,0.74}
\usepackage[pagebackref,breaklinks,colorlinks,allcolors=cvprblue,pdftitle={Four Tokens to Find Waldo: Steering Deterministic Visual Search in Frozen VLMs},pdfauthor={Feng Wang, Yaojie Liu, Zeeshan Hayder, Vu Minh Hieu Phan, Alan Yuille, Wen-Sheng Chu}]{hyperref}
\usepackage{booktabs}
\usepackage{url}
\usepackage{graphicx}
\usepackage{multirow}
\usepackage{amsmath}
\usepackage{mathtools}

\providecolor{darkgreen}{rgb}{0.0, 0.5, 0.0}

\title{Four Tokens to Find Waldo:\\ Steering Deterministic Visual Search in Frozen VLMs}

\makeatletter
\newcommand\blfootnote[2]{%
  \begingroup
  \def\@thefnmark{#1}%
  \NoHyper\@footnotetext{#2}\endNoHyper
  \endgroup
}
\makeatother

\author{
Feng Wang$^{1,2*\dagger}$\hspace{0.40em}Yaojie Liu$^{1*}$\hspace{0.40em}Zeeshan Hayder$^{1}$\hspace{0.40em}Vu Minh Hieu Phan$^{1}$\hspace{0.40em}Alan Yuille$^{2}$\hspace{0.40em}Wen-Sheng Chu$^{1}$\\[3pt]
$^{1}$Google\qquad $^{2}$Johns Hopkins University\\[3pt]
{\normalsize\href{https://4twaldo.github.io}{\texttt{https://4twaldo.github.io}}}
}

\begin{document}

\maketitle

\begin{strip}
    \vspace{-15.5mm}
    \centering
    \includegraphics[width=1\textwidth]{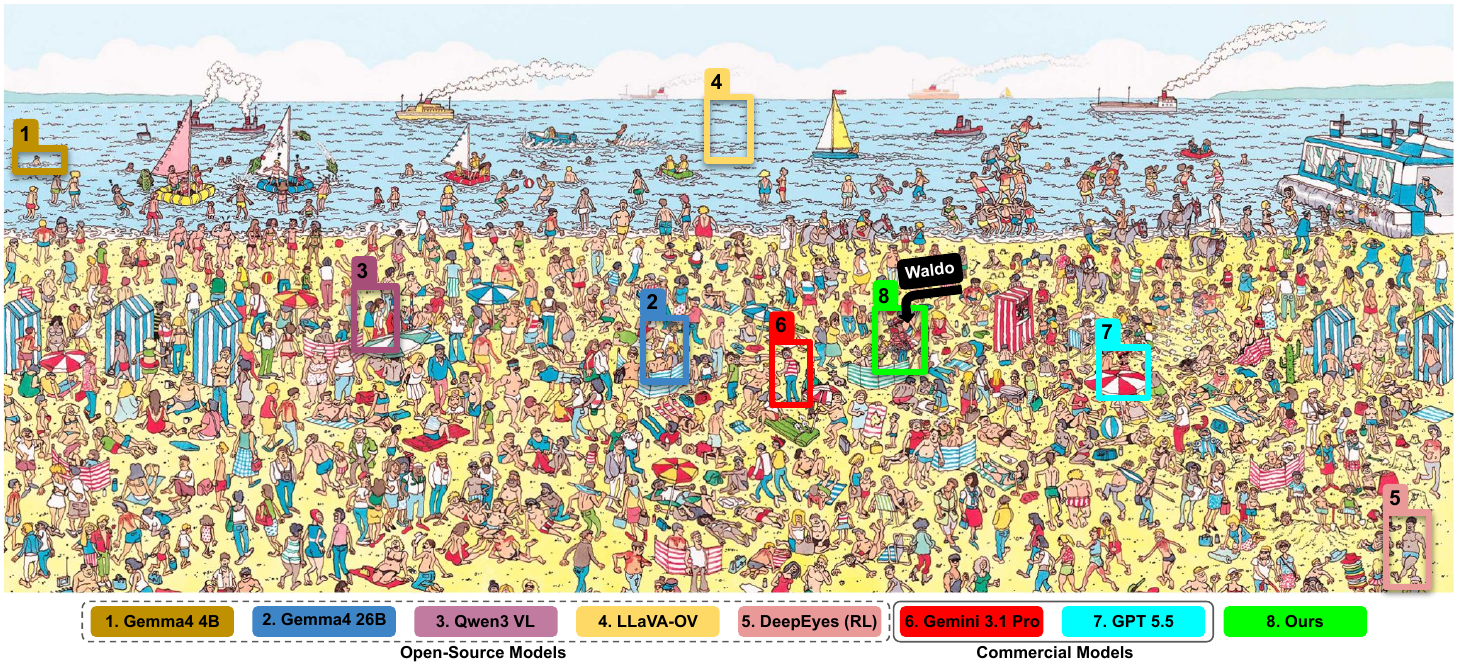}
    \vspace{-4ex}
    \captionof{figure}[Teaser]{\small\textit{\textbf{Where is Waldo?}} A demonstration of the fine-grained visual reasoning bottleneck in modern VLMs. When searching crowded scenes, standard models lose fine details during downsampling, while open-ended multi-turn methods like DeepEyes still miss the target. By learning just four prefix tokens to steer a frozen VLM through a deterministic three-turn search, our method accurately localizes Waldo (image resolution: $2048\times 834$ pixels; best viewed under magnification).}
    \label{fig:waldo_teaser}
    \vspace{-2mm}
\end{strip}

\blfootnote{$*$}{Equal contribution.}
\blfootnote{$\dagger$}{This work was done during Feng Wang's internship at Google.}

\begin{abstract}
To tackle fine-grained visual reasoning, such as the classic ``Where is Waldo?'', recent Vision-Language Models (VLMs) adopt active visual search, which dynamically examines images in a multi-turn paradigm.
Existing methods execute the search as a sequential chain of crop-and-zoom actions and reasoning across turns, resulting in an open-ended trajectory that requires costly full-backbone reinforcement learning (RL) and incurs high latency.
In fact, modern VLMs can already reason over the global scene and locate regions of interest in a single turn, allowing them to be analyzed independently for fine local details rather than through step-by-step exploration.
Building on this insight, we collapse open-ended search into a deterministic three-turn trajectory: 
the VLM first proposes a search policy and all candidate crops from the global view, next inspects each zoomed-in patch in parallel to produce local findings, and then aggregates them into a final answer.
Fully leveraging the VLM's own capabilities, we can steer this search simply by learning as few as four prefix tokens via supervised fine-tuning while keeping the model completely frozen.
Across five challenging fine-grained benchmarks, our method consistently outperforms both multi-turn RL methods and frontier proprietary models, surpassing Claude Opus 4.8 by 15.3\% on WaldoBench and boosting Gemini 3.1 Pro by up to 50.4\% when transferring proposed crops as a plug-and-play visual front-end.

\end{abstract}

\vspace{-2ex}
\section{Introduction}
\vspace{-1ex}

\begin{figure*}[t]
    \centering
    \includegraphics[width=0.9\textwidth]{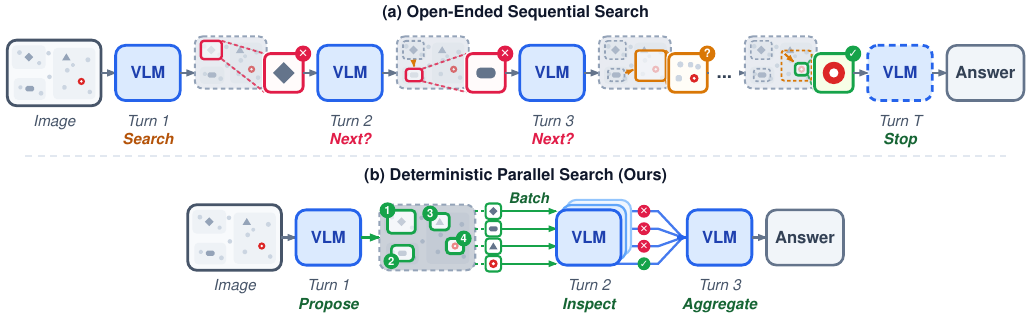}
    \vspace{-1ex}
    \caption{\textbf{Comparison of active visual search paradigms.} 
    \textbf{(a) Open-Ended Sequential Search} searches one step at a time over a variable horizon ($T$ turns), moving across missed regions (\redcross) and refining coarse crops (\orangequestion) before isolating the target (\greencheck), which incurs high latency and requires costly full-backbone RL. 
    \textbf{(b) Deterministic Parallel Search (Ours)} collapses open-ended exploration into a fixed three-turn trajectory: the VLM proposes all candidate regions at tailored zoom-in scales in Turn~1 (\emph{Propose}), inspects the resulting crops as a parallel batch in Turn~2 (\emph{Inspect}), and aggregates local findings into the final answer in Turn~3 (\emph{Aggregate}).}
    \label{fig:paradigm_comparison}
    \vspace{-1ex}
\end{figure*}

Consider the classic \textit{``Where is Waldo?''} puzzle in~\cref{fig:waldo_teaser}: how does a human observer locate a tiny target in such a crowded scene? 
Rather than resolving every detail in a single glance, human vision follows a coarse-to-fine search process, scanning the global scene for candidate areas before zooming in to inspect local regions. 
This ability to actively search and reason over localized visual cues underpins fine-grained visual reasoning, from small-object grounding~\cite{wu2024visualsearch,wang2024hrbench} and dense counting~\cite{chattopadhyay2017counting,ranjan2021learning,paiss2023teaching} to spotting subtle image differences~\cite{jhamtani2018learning,park2019robust}. 
Mastering these tasks is critical as Vision-Language Models (VLMs)~\cite{liu2023visual,achiam2023gpt4,reid2024gemini,team2024gemma,bai2025qwen2_5_vl} are deployed to navigate user interfaces~\cite{hong2024cogagent,cheng2024seeclick}, read dense documents~\cite{mathew2021docvqa}, and monitor real-world environments. 
Yet, despite strong holistic perception, leading open-source and commercial VLMs still fail on these fine-grained tasks, as illustrated in~\cref{fig:waldo_teaser}.

When encoding a high-resolution image in a single pass, standard VLMs downsample it to a bounded token budget, obscuring fine details. 
To address this, recent works adopt \textbf{active visual search}, enabling VLMs to perform multi-turn inspection across local image regions. 
Early approaches~\cite{hu2024visualsketchpad, li2025dyfo, shen2024zoomeye, wu2024visualsearch} rely on external vision models, such as Grounding DINO~\cite{liu2024groundingdino} or SAM~\cite{kirillov2023sam}, to locate candidate regions, while recent multi-turn reinforcement learning (RL) methods~\cite{chng2026sensesearch,hong2025deepeyesv2,hou2026codev,liang2026adaturn,su2025pixelreasoner,zheng2025deepeyes} internalize the search policy within a single VLM. 
As shown in~\cref{fig:paradigm_comparison}a, previous methods organize visual search as a sequential trajectory of crop-and-zoom actions, analogous to how a single turn rolls out chain-of-thought (CoT) reasoning: at each turn, the system reasons over the current view and either zooms into a new crop for the next turn or stops to answer. Exploring one region per turn makes this search trajectory open-ended, where both the visited crops and the search horizon $T$ vary with the image content. At inference time, this sequential search incurs high latency; during training, optimizing such trial-and-error exploration across variable turns requires large-scale rollouts and costly full-backbone RL.

Treating the search trajectory as another CoT process, however, is largely redundant: a single global view already enables modern VLMs to understand scene layout and decompose the search into regional sub-tasks (\eg, verifying a target object or counting objects in an area). What this global view lacks is primarily the fine detail required to answer each sub-task; without sequential dependencies across candidate areas, these sub-tasks can be distributed and executed concurrently rather than one turn at a time. Motivated by this observation, we reformulate active visual search into a \textbf{deterministic} three-turn search trajectory, as shown in~\cref{fig:paradigm_comparison}b. In Turn~1 (\emph{Propose}), given the global image and query, the VLM outlines these regional sub-tasks and writes Python code specifying the crop coordinates and zoom-in factors for all candidate regions at once. In Turn~2 (\emph{Inspect}), each re-encoded high-resolution patch is paired with the Turn~1 global input and search plan and examined in a parallel batch to complete its assigned sub-task; in Turn~3 (\emph{Aggregate}), the model gathers the regional results to reconcile overlaps and produce the final prediction.

Collapsing open-ended exploration into this deterministic trajectory also changes how the search policy can be learned. Within this structure, each turn builds on capabilities that modern VLMs~\cite{team2024gemma,bai2025qwen2_5_vl} possess from pre-training, including visual perception and grounding, task decomposition, code generation, and contextual reasoning. Learning the search policy therefore reduces to coordinating these capabilities across the three turns, which we achieve by prepending as few as {\em four} learnable prefix tokens to the Turn~1 input (shared across subsequent turns via the KV-cache) and optimizing them via supervised fine-tuning (SFT) on a completely frozen backbone.
Evaluated on WaldoBench~\cite{waldobench2026}, $V^*$~Bench~\cite{wu2024visualsearch}, UNICBench~\cite{unicbench2026}, HR-Bench~\cite{wang2024hrbench}, and VDiff-Bench-Gen~\cite{vdiffbench2026}, our method surpasses both recent multi-turn RL methods and proprietary models such as Claude Opus 4.8 and Gemini 3.1 Pro, while serving as a plug-and-play visual front-end that boosts downstream VLM accuracy on challenging fine-grained visual tasks
by 50.4\%--54.6\%.
In summary, our contributions are threefold:
\begin{itemize}
    \item We reformulate active visual search into a deterministic three-turn trajectory that \emph{proposes} regional sub-tasks, \emph{inspects} crops in parallel, and \emph{aggregates} results.
    \item We show that visual search only requires coordinating native VLM capabilities across turns, steering a completely frozen model with as few as four prefix tokens.
    \item Across five benchmarks, we surpass multi-turn RL and frontier models, while boosting off-the-shelf VLMs by up to 54.6\% via zero-shot crop transfer.
\end{itemize}

\section{Related Work}
\Paragraph{High-Resolution Processing via Image Slicing.}
To bypass fixed-resolution vision encoders~\cite{radford2021clip,zhai2023siglip}, modern VLMs adopt dual-encoder designs~\cite{li2024minigemini,luo2025feast} or grid-slicing schemes such as AnyRes and dynamic patchification~\cite{chen2024internvl,li2024llavaonevision,li2024monkey,lin2024sphinx,liu2024llavanext,guo2024llavauhd,dong2024internlm_xcomposer2_4khd,yao2024minicpm_v}, which split an image into uniform sub-tiles encoded alongside a global thumbnail. While effective for documents and mid-sized objects, uniform grids waste visual tokens and KV-cache memory on background regions, and arbitrary tile seams often bisect small targets. Adaptive tiling and token pruning~\cite{jiang2025teva,chen2024fastv,shang2024llava_prumerge} drop uninformative patches, but remain constrained to coarse predefined grids, whereas our method predicts query-tailored free-form crops and zoom-in factors.

\Paragraph{Active Visual Search.}
Rather than encoding static grids, active visual search dynamically zooms into task-relevant regions. Early visual programming~\cite{gupta2023visual,suris2023vipergpt} generates single-pass code from text to invoke external vision models, while multi-turn frameworks search outside the VLM via specialist detectors~\cite{hu2024visualsketchpad}, auxiliary search networks ($V^*$~\cite{wu2024visualsearch}, CropVLM~\cite{carvalho2026cropvlm}), or heuristic tree search (ZoomEye~\cite{shen2024zoomeye}, DyFo~\cite{li2025dyfo}), remaining bounded by external detector capabilities or rigid search hierarchies. Recent multi-turn VLMs internalize this search policy via supervised fine-tuning~\cite{qi2024cogcom,shao2024visual_cot} or reinforcement learning with group/sequence policy optimization~\cite{shao2024deepseekmath,chng2026sensesearch,hong2025deepeyesv2,zheng2025deepeyes}, curiosity-driven rewards~\cite{su2025pixelreasoner}, tool-aware RL~\cite{hou2026codev}, and budget-aware scaling~\cite{liang2026adaturn}. Even so, their open-ended sequential trajectories incur high latency, suffer from drift or truncation on dense counting, and demand costly full-backbone RL with sparse credit assignment. In contrast, we combine model-free Python cropping with a deterministic three-turn trajectory (\emph{Propose}, \emph{Inspect}, \emph{Aggregate}), executing regional sub-tasks in parallel and steering a frozen VLM via SFT on as few as four prefix tokens.

\Paragraph{Attention Guidance and Latent Visual Reasoning.}
Another line of work targets fine-grained perception without high-resolution cropping, via internal attention guidance~\cite{yu2024api,liu2024pai,zou2025memvr}, referential grounding~\cite{chen2023shikra,lai2024lisa,tian2024glamm,yuan2024osprey,peng2023kosmos2,you2023ferret}, or latent visual reasoning~\cite{li2026lvr,li2026livr,yang2025mirage}. These methods align cross-attention or special tokens with object boundaries, or deliberate internally over latent visual features before answering. While avoiding extra encoding cost, they operate strictly on the initial downsampled tokens and cannot recover local details already lost during vision encoding. Our approach instead uses lightweight prefix tokens~\cite{lester2021power,li2021prefix,jia2022vpt} on a frozen VLM to trigger physical high-resolution cropping and re-encoding across three turns.

\section{Methodology}
\label{sec:method}

\begin{figure*}[t]
    \centering
    \includegraphics[width=1\textwidth]{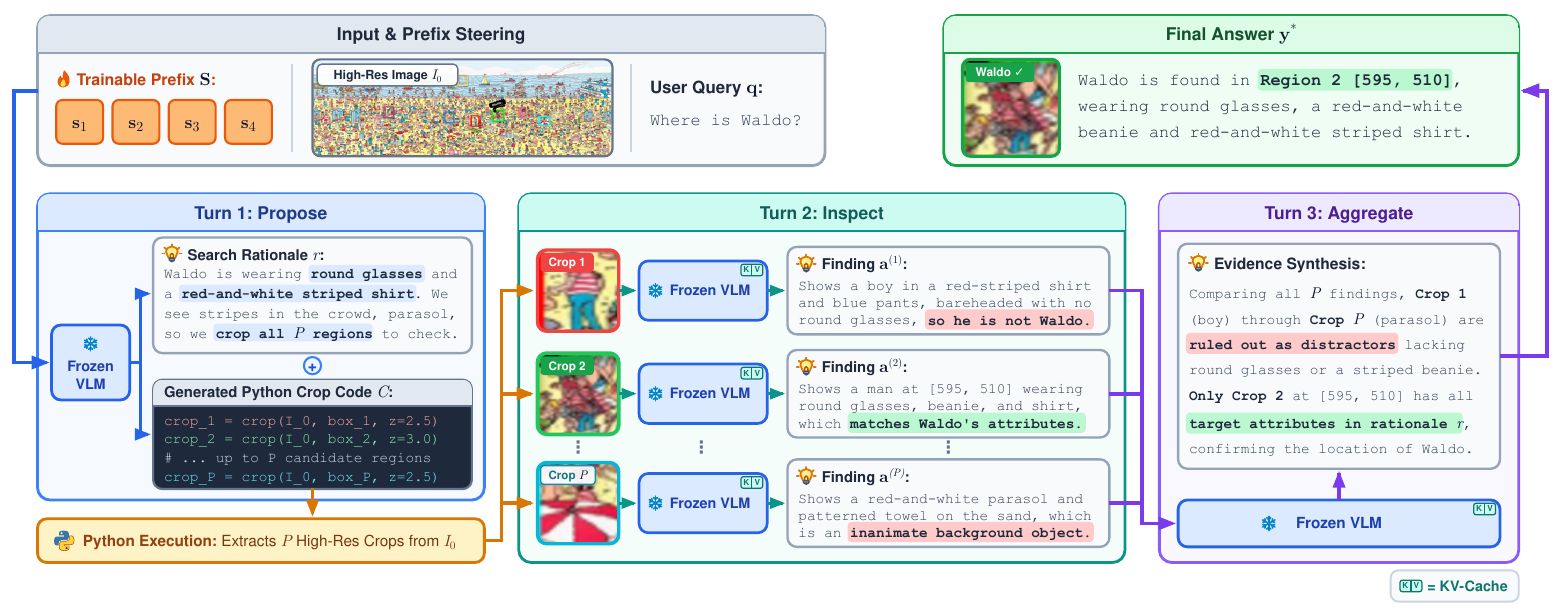}
    \vspace{-5ex}
    \caption{\small\textbf{Overview of our deterministic three-turn visual search framework.} Steered by trainable prefix tokens $\mathbf{S}$ prepended to the image $I_0$ and query $\mathbf{q}$, a frozen VLM generates a search rationale $r$ (outlining the global analysis and instructions for all three turns) and Python crop code $\mathcal{C}$ in \textbf{Turn~1 (\emph{Propose})}, inspects the $P$ extracted crops in parallel to produce regional findings $\{\mathbf{a}^{(p)}\}_{p=1}^P$ in \textbf{Turn~2 (\emph{Inspect})}, and aggregates the regional evidence into the final answer $\mathbf{y}^*$ in \textbf{Turn~3 (\emph{Aggregate})}. The KV-Cache icon indicates reusing Turn~1's cached states across all parallel branches in Turn~2, and reusing both Turn~1 and Turn~2 cached states in Turn~3.}
    \label{fig:method_overview}
\end{figure*}

\subsection{Preliminaries}
\label{sec:preliminaries}

Let $I_0 \!\in\! \mathbb{R}^{H_0 \times W_0 \times 3}$ denote a high-resolution input image (or image pair for comparison) and $\mathbf{q}$ be a text query for fine-grained visual reasoning. 
In conventional VLMs, a vision encoder $\mathcal{E}_v$ maps $I_0$ into a downsampled sequence of $D$-dimensional visual tokens $\mathbf{V}_0 = \mathcal{E}_v(I_0) \in \mathbb{R}^{N_v \times D}$ with a bounded token budget $N_v \ll H_0 \times W_0$. Given $(\mathbf{V}_0, \mathbf{q})$, a single-turn VLM parameterized by $\Theta$ directly generates the answer sequence $\mathbf{y} \!=\! (y_1, \dots, y_L)$ via its autoregressive policy $\pi_\Theta$:
\begin{equation}
    \pi_\Theta\left(\mathbf{y} \mid \mathbf{V}_0, \mathbf{q}\right) = \prod_{l=1}^{L} p_\Theta\left(y_l \;\middle|\; \mathbf{V}_0, \mathbf{q}, \mathbf{y}_{<l}\right).
    \label{eq:single_turn_vlm}
\end{equation}
With bounded $N_v$, small targets and local details are heavily compressed in $\mathbf{V}_0$, creating a resolution bottleneck.
To address this, existing active visual search methods iteratively extend~\cref{eq:single_turn_vlm} across $T$ turns before predicting $\mathbf{y}$:
\begin{align}
    p_\Theta(\mathbf{y}, \mathbf{h}_{1:T} \mid I_0, \mathbf{q}) = & \underbrace{\left[\prod_{t=1}^{T} \pi_\Theta\left(r_t, \mathcal{C}_t \;\middle|\; \mathbf{V}_0, \mathbf{q}, \mathbf{h}_{<t}\right)\right]}_{\text{Sequential region cropping ($T$ turns)}} \nonumber \\
    & \cdot \underbrace{\pi_\Theta\left(\mathbf{y} \;\middle|\; \mathbf{V}_0, \mathbf{q}, \mathbf{h}_{1:T}\right)}_{\text{Final answer generation}},
    \label{eq:open_ended_search}
\end{align}
where each turn $t$ generates a reasoning rationale $r_t$ and cropping action $\mathcal{C}_t$, appending $(r_t, \mathcal{C}_t, \mathbf{V}_t)$ to the turn history $\mathbf{h}_{<t}$ with $\mathbf{V}_t \!=\! \mathcal{E}_v \left(\mathrm{Exec}(\mathcal{C}_t, I_0)\right)$ and $\mathrm{Exec}(\cdot)$ executing $\mathcal{C}_t$ to crop $I_0$. While $\mathbf{h}_{1:T}$ restores local resolution, conditioning each step on $\mathbf{h}_{<t}$ yields an open-ended trajectory whose visited regions and horizon $T$ vary per image, incurring high latency and complicating policy optimization.

\subsection{Deterministic Three-Turn Visual Search}
\label{sec:three_turn_search}
Although $\mathbf{V}_0$ lacks fine local detail, a single pass over its global view already allows a modern VLM to parse the scene layout and decompose the search into regional sub-tasks. Without sequential dependencies across candidate areas, these sub-tasks can be executed concurrently rather than chained across turns. Motivated by this, we collapse the open-ended $T$-turn chain in~\cref{eq:open_ended_search} into a \textbf{deterministic} three-turn trajectory $\tau$ with a fixed stage topology ($T=3$), while allowing the candidate count $P$, crop coordinates, and zoom-in scales to adapt dynamically per query. Letting $\pi_\Theta^{(1)}, \pi_\Theta^{(2)}, \pi_\Theta^{(3)}$ denote the VLM policy $\pi_\Theta$ at each turn (\cref{fig:method_overview}), the search proceeds as follows:

\vspace{0.5ex}\noindent
{\bf Turn 1 (\emph{Propose}).}
Given an input image $I_0$ and a text query $\mathbf{q}$, the VLM externalizes its search policy as $(r, \mathcal{C})$, comprising a search rationale $r$ (which analyzes the global layout and specifies sub-task instructions for downstream inspection and aggregation) and an executable Python script $\mathcal{C}$ specifying all $P$ candidate regions at once:
\begin{equation}
    (r, \mathcal{C}) \sim \pi_\Theta^{(1)}\left(\cdot \;\middle|\; \mathbf{V}_0, \mathbf{q}\right).
    \label{eq:turn1_propose}
\end{equation}
In code, $\mathcal{C}$ specifies normalized crop boxes and query-tailored zoom-in factors $z_p$ on $I_0$, supporting both targeted multi-scale region-of-interest (RoI) cropping and programmatic grid partitions for dense scenes:
\begin{equation}
    \left\{I^{(1)}, I^{(2)}, \dots, I^{(P)}\right\} = \mathrm{Exec}\left(\mathcal{C}, I_0\right),
    \label{eq:python_exec}
\end{equation}
where each extracted patch $I^{(p)}$ is cropped directly from native $I_0$ and rescaled by $z_p$ to preserve fine local fidelity. At the end of Turn~1, the KV states of the sequence $(\mathbf{V}_0, \mathbf{q}, r, \mathcal{C})$ are cached once in the KV-cache to be reused across subsequent turns. 

\vspace{0.5ex}\noindent
{\bf Turn 2 (\emph{Inspect}).}
The execution environment encodes each crop into $\mathbf{V}^{(p)} \!=\! \mathcal{E}_v(I^{(p)})$. Instead of concatenating all $P$ crops into one long prompt or scanning them sequentially across turns, the model evaluates the candidate regions independently in parallel as a batch of $P$ branches ($p \in \{1, \dots, P\}$). Sharing the prefilled Turn~1 KV-cache without recomputing $(\mathbf{V}_0, \mathbf{q}, r, \mathcal{C})$, each branch attends to its assigned patch $\mathbf{V}^{(p)}$ and follows the instructions in $r$ from the Turn~1 context to resolve its sub-task and generate a concise regional finding $\mathbf{a}^{(p)}$ (\eg, target verification, sub-counts, or local coordinates):
\begin{equation}
    \mathbf{a}^{(p)} \sim \pi_\Theta^{(2)}\left(\cdot \;\middle|\; \mathbf{V}_0, \mathbf{q}, r, \mathcal{C}, \mathbf{V}^{(p)}\right).
    \label{eq:turn2_inspect}
\end{equation}
Conditioning each branch on a single patch $\mathbf{V}^{(p)}$ prevents cross-patch attention interference while computing and caching the branch KV states for $(\mathbf{V}^{(p)}, \mathbf{a}^{(p)})$ in a single batched pass.

\vspace{0.5ex}\noindent
{\bf Turn 3 (\emph{Aggregate}).}
To synthesize the final prediction, Turn~3 assembles the shared KV-cache from Turn~1 together with all $P$ cached branch segments $\{\mathbf{V}^{(p)}, \mathbf{a}^{(p)}\}_{p=1}^P$ from Turn~2. Following the aggregation instructions in $r$ over the global context, high-resolution patches, and regional findings without re-encoding or re-prefilling prior turns, the model aggregates the local evidence (\eg, filtering out distractors, reconciling boundary overlaps, or removing duplicate counts) to generate the final answer $\mathbf{y}^*$:
\begin{equation}
    \mathbf{y}^* \sim \pi_\Theta^{(3)}\left(\cdot \;\middle|\; \mathbf{V}_0, \mathbf{q}, r, \mathcal{C}, \{\mathbf{V}^{(p)}, \mathbf{a}^{(p)}\}_{p=1}^P\right).
    \label{eq:turn3_aggregate}
\end{equation}
Combining \cref{eq:turn1_propose,eq:turn2_inspect,eq:turn3_aggregate} (omitting turn contexts for brevity), the joint probability of our three-turn trajectory $\tau \!=\! (r, \mathcal{C}, \{\mathbf{a}^{(p)}\}_{p=1}^P, \mathbf{y}^*)$ factorizes into three deterministic stages:
\begin{align}
    & p_\Theta(\tau \mid I_0, \mathbf{q}) \nonumber\\
    & \quad = \underbrace{\pi_\Theta^{(1)}(r, \mathcal{C})}_{\text{Turn 1 (\emph{Propose})}} \cdot \underbrace{\left[\prod_{p=1}^{P} \pi_\Theta^{(2)}(\mathbf{a}^{(p)})\right]}_{\text{Turn 2 (\emph{Inspect})}} 
    \cdot \underbrace{\pi_\Theta^{(3)}(\mathbf{y}^*)}_{\text{Turn 3 (\emph{Aggregate})}}.
    \label{eq:three_turn_joint}
\end{align}

\subsection{Policy Steering via Learnable Prefix Tokens}
\label{sec:prefix_sft}
Given an input image and query $(\mathbf{V}_0, \mathbf{q})$, an off-the-shelf VLM defaults to single-pass answering (\cref{eq:single_turn_vlm}). 
Yet, modern VLMs~\cite{team2024gemma,bai2025qwen2_5_vl} already possess the capabilities needed for our three-turn search, including task decomposition, code generation, and visual reasoning, from pre-training. 
Therefore, we only need a control signal to switch the VLM out of single-pass answering and direct it through the three-turn search policy. Text prompting alone is insufficient, as discrete vocabulary tokens are too weak to override the model's single-turn alignment prior across turns. Instead, keeping all backbone weights $\Theta$ frozen, we prepend $K$ learnable continuous prefix tokens $\mathbf{S} = [\mathbf{s}_1, \mathbf{s}_2, \dots, \mathbf{s}_K] \in \mathbb{R}^{K \times D}$ ($K \ll N_v$) to the Turn~1 input:
\begin{equation}
    (\mathbf{V}_0, \mathbf{q}) \;\longrightarrow\; (\mathbf{S}, \mathbf{V}_0, \mathbf{q}).
    \label{eq:prefix_concat}
\end{equation}
Operating in continuous embedding space, $\mathbf{S}$ projects into $K$ persistent key-value pairs at every self-attention layer. Anchored at the front of the shared KV-cache across Turns~1--3 (\cref{eq:prefix_concat}), these vectors act as learned attention biases directing frozen attention heads to coordinate \emph{Propose}, \emph{Inspect}, and \emph{Aggregate} without turn-specific parameters.

{\bf Supervised Trajectory Optimization.}
Let $\mathcal{D}$ be a curated dataset of reference three-turn search trajectories  $\{(I_0^{(i)}, \mathbf{q}^{(i)}, \tau^{(i)})\}_{i=1}^{|\mathcal{D}|}$ (see Appendix~B for data curation details).
During training, we optimize exclusively over the prefix tokens $\mathbf{S}$ on frozen $\Theta$ by minimizing the token-level negative log-likelihood of the factorized trajectory distribution in \cref{eq:three_turn_joint}:
\begin{equation}
    \mathcal{L}(\mathbf{S}) = -\sum_{i=1}^{|\mathcal{D}|} \log p_\Theta\left(\tau^{(i)} \;\middle|\; I_0^{(i)}, \mathbf{q}^{(i)}, \mathbf{S}\right).
    \label{eq:sft_loss}
\end{equation}
With three-turn factorization fixed across all examples, supervised fine-tuning provides dense token-level supervision at every turn in a single offline reference trajectory, avoiding online multi-turn RL rollouts and sparse credit assignment while preserving the frozen backbone's capabilities.

\section{Experiments}
\label{sec:experiments}

\subsection{Setup}
\label{subsec:exp_setup}

{\bf Benchmarks.}
We evaluate on 5 fine-grained benchmarks of increasing complexity: 
(1) {\bf WaldoBench}~\cite{waldobench2026} for localizing a single object in extreme clutter,
(2) {\bf $V^*$~Bench}~\cite{wu2024visualsearch} for reasoning about attributes and spatial relationships of small objects,
(3) {\bf UNICBench}~\cite{unicbench2026} for counting in dense multi-object scenes, 
(4) {\bf HR-Bench}~\cite{wang2024hrbench} for perceiving fine detail in 4K/8K images, and
(5) {\bf VDiff-Bench-Gen}~\cite{vdiffbench2026} for detecting discrepancies across multiple images.

{\bf Implementation Details.}
We evaluate our framework on frozen Gemma-4-4B and Gemma-4-26B~\cite{team2024gemma} backbones, setting the default prefix length to $K \!=\! 4$ (\eg, $4 \times 2560 \approx 10.2\text{K}$ trainable parameters, or 20.5KB in BF16 on Gemma-4-4B). We train $\mathbf{S}$ via supervised fine-tuning using the AdamW optimizer~\cite{loshchilov2019adamw} ($\beta_1 = 0.9, \beta_2 = 0.99$) with a weight decay of 0.05. The learning rate is initialized at $2 \!\times\! 10^{-5}$ and decayed to $1 \!\times\! 10^{-6}$ via a cosine annealing schedule. In Turn~1, crop coordinates are predicted and executed in Python on the native high-resolution image $I_0$. All models are implemented in PyTorch~\cite{paszke2017automatic} with FlashAttention-2~\cite{dao2023flashattention2} and trained on 8 NVIDIA A100 GPUs using fully sharded data parallel.

\begin{table}[t]
  \centering
  \small
  \setlength{\tabcolsep}{4pt}
  \caption{\textbf{Visual Search in Extreme Clutter:} Localization accuracy on WaldoBench~\citep{waldobench2026}. Baseline metrics are sourced from the official leaderboard~\citep{waldobench2026} (v0.0.3).}
  \vspace{-1mm}
  \label{tab:waldobench}
  \begin{tabularx}{\linewidth}{Xc}
    \toprule
    \textbf{Model} & \textbf{Accuracy (\%)} \\
    \midrule
    \multicolumn{2}{l}{\textcolor{slateheader}{\textit{Proprietary Frontier Multimodal Models}}} \\
    Gemini 3.1 Pro & 26.4 \\
    Claude Opus 4.8 & \underline{37.7} \\
    GPT-5.5 & 35.8 \\
    \midrule
    \multicolumn{2}{l}{\textcolor{slateheader}{\textit{Open-Weight Direct-Reasoning VLMs}}} \\
    Qwen2.5-VL-32B~\citep{bai2025qwen2_5_vl} & 0.0 \\
    Gemma-4-26B~\citep{team2024gemma} & 0.0 \\
    \midrule
    \rowcolor{oursbg}\textbf{Ours (Gemma-4-26B)} & \textbf{53.0} \\
    \bottomrule
  \end{tabularx}
  \vspace{-1mm}
\end{table}

\begin{table*}[thb!]
\centering
\small
\renewcommand{\arraystretch}{0.88}
\setlength{\tabcolsep}{6pt}
\caption{\textbf{Fine-Grained Grounding and High-Resolution Scaling:} Accuracy comparison (\%) on $V^*$~Bench~\citep{wu2024visualsearch}, HR-Bench 4K, and HR-Bench 8K~\citep{wang2024hrbench}. \textit{Attr.} and \textit{Spat.} denote Attribute Recognition and Spatial Reasoning on $V^*$~Bench; \textit{FSP} and \textit{FCP} denote Fine-grained Single-instance and Cross-instance Perception on HR-Bench.}
\vspace{-2mm}
\label{tab:vstar_results}
\label{tab:hrbench_4k_results}
\label{tab:hrbench_8k_results}
\begin{tabularx}{\textwidth}{X ccc ccc ccc}
\toprule
& \multicolumn{3}{c}{\textbf{$V^*$~Bench}} & \multicolumn{3}{c}{\textbf{HR-Bench 4K}} & \multicolumn{3}{c}{\textbf{HR-Bench 8K}} \\
\cmidrule(lr){2-4} \cmidrule(lr){5-7} \cmidrule(lr){8-10}
\textbf{Model} & \textbf{Attr.} & \textbf{Spat.} & \textbf{Overall} & \textbf{FSP} & \textbf{FCP} & \textbf{Overall} & \textbf{FSP} & \textbf{FCP} & \textbf{Overall} \\
\midrule
\multicolumn{10}{l}{\textcolor{slateheader}{\textit{General / Direct-Reasoning VLMs}}} \\
GPT-4o~\citep{achiam2023gpt4} & -- & -- & 66.0 & 70.0 & 48.0 & 59.0 & 62.0 & 49.0 & 55.5 \\
LLaVA-OneVision (7B)~\citep{li2024llavaonevision} & 75.7 & 75.0 & 75.4 & 72.0 & 54.0 & 63.0 & 67.3 & 52.3 & 59.8 \\
Qwen2.5-VL (7B)~\citep{bai2025qwen2_5_vl} & 73.9 & 67.1 & 71.2 & 85.2 & 52.2 & 68.8 & 78.8 & 51.8 & 65.3 \\
Qwen2.5-VL (32B)~\citep{bai2025qwen2_5_vl} & 87.8 & 88.1 & 87.9 & 89.8 & 58.0 & 73.9 & 84.5 & 56.3 & 70.4 \\
\midrule
\multicolumn{10}{l}{\textcolor{slateheader}{\textit{Active Visual Search Methods}}} \\
SEAL~\citep{wu2024visualsearch} & 74.8 & 76.3 & 75.4 & -- & -- & -- & -- & -- & -- \\
ZoomEye~\citep{shen2024zoomeye} & \underline{93.9} & 85.5 & \underline{90.6} & 84.3 & 55.0 & 69.6 & \underline{88.5} & 50.0 & 69.3 \\
Pixel-Reasoner (7B)~\citep{su2025pixelreasoner} & 83.5 & 76.3 & 80.6 & 86.0 & 60.3 & 72.9 & 80.0 & 54.3 & 66.9 \\
DeepEyes (7B)~\citep{zheng2025deepeyes} & 91.3 & \underline{88.2} & 90.1 & \underline{91.3} & 59.0 & 75.1 & 86.8 & 58.5 & 72.6 \\
DeepEyesV2 (7B)~\citep{hong2025deepeyesv2} & -- & -- & 81.8 & -- & -- & \underline{77.9} & -- & -- & \underline{73.8} \\
CodeV (7B)~\citep{hou2026codev} & -- & -- & 84.8 & 91.0 & \underline{61.3} & 76.1 & 81.3 & \underline{60.3} & 71.3 \\
\midrule
\rowcolor{oursbg}\textbf{Ours (Gemma-4-4B)} & \textbf{94.2} & \textbf{88.5} & \textbf{92.0} & \textbf{95.5} & \textbf{68.7} & \textbf{82.1} & \textbf{92.4} & \textbf{66.4} & \textbf{79.4} \\
\bottomrule
\end{tabularx}
\vspace{-1mm}
\end{table*}

\subsection{Results}
\label{subsec:main_results}

\noindent
{\bf Visual Search in Clutter on WaldoBench \cite{waldobench2026}.}
\label{subsec:waldobench}
We first evaluate localizing Waldo in extreme clutter, where the target tends to occupy $\le\!0.1\%$ of the image area and accuracy measures whether the predicted 2D coordinate falls within $5\%$ of the ground-truth location (\cref{tab:waldobench}). Since fixed-resolution encoding blurs tiny targets during global downsampling, direct-reasoning open-weight models fail completely, with both Qwen2.5-VL-32B~\cite{bai2025qwen2_5_vl} and Gemma-4-26B~\cite{team2024gemma} scoring \textbf{0.0\%}.
Proprietary models with automated zooming perform better (Claude Opus 4.8 at 37.7\%, GPT-5.5 at 35.8\%, Gemini 3.1 Pro at 26.4\%), yet their heuristic exploration frequently stalls among look-alike distractors. Steered by four prefix tokens on the same frozen Gemma-4-26B backbone, our method lifts accuracy from 0.0\% to \textbf{53.0\%}, surpassing the best-performing Claude Opus 4.8 by \textbf{+15.3\%}.

\noindent
{\bf Tiny Object Grounding/Perception on \texorpdfstring{$V^*$}{V*}~Bench \cite{wu2024visualsearch}.}
\label{subsec:grounding}
We next evaluate attribute recognition and spatial reasoning over small objects in natural scenes on $V^*$~Bench~\cite{wu2024visualsearch}, measured by multiple-choice VQA accuracy (\cref{tab:vstar_results}). Direct-reasoning VLMs struggle with local details, with GPT-4o~\cite{achiam2023gpt4} reaching 66.0\% overall accuracy and 7B open-weight models~\cite{li2024llavaonevision,bai2025qwen2_5_vl} achieving 71.2\%--75.4\%. 
Active visual search methods such as ZoomEye~\cite{shen2024zoomeye} (90.6\%), DeepEyes-7B~\cite{zheng2025deepeyes} (90.1\%), and CodeV-7B~\cite{hou2026codev} (84.8\%) improve localization accuracy through multi-turn sequential exploration. Using a smaller frozen Gemma-4-4B~\cite{team2024gemma} backbone, our deterministic three-turn search reaches \textbf{94.2\%} on Attribute, \textbf{88.5\%} on Spatial, and \textbf{92.0\%} overall, outperforming all alternative 7B sequential search methods.

\begin{table}[t]
\centering
\small
\setlength{\tabcolsep}{3.5pt}
\caption{\textbf{Massive Object Counting:} Accuracy (\%) on UNICBench~\citep{unicbench2026} across difficulty tiers. Our method substantially outperforms foundation models, programmatic baselines, and latent visual thinking, particularly on the dense \textit{Hard} split.}
\vspace{-1ex}
\label{tab:unicbench_results}
\begin{tabularx}{\linewidth}{Xccc}
\toprule
\textbf{Model} & \textbf{Easy} & \textbf{Medium} & \textbf{Hard} \\
\midrule
\multicolumn{4}{l}{\textcolor{slateheader}{\textit{General Multimodal Foundation Models}}} \\
Gemma-4-4B~\citep{team2024gemma} & 74.4 & 27.2 & 9.9 \\
Gemma-4-26B~\citep{team2024gemma} & 76.8 & 27.5 & 10.1 \\
Gemini 3.1 Pro~\citep{reid2024gemini} & \underline{85.5} & \underline{34.0} & \underline{15.8} \\
\midrule
\multicolumn{4}{l}{\textcolor{slateheader}{\textit{Visual Programming Baselines}}} \\
ViperGPT~\citep{suris2023vipergpt} & 84.5 & 21.0 & 5.2 \\
\midrule
\multicolumn{4}{l}{\textcolor{slateheader}{\textit{Latent Visual Reasoning Baselines}}} \\
Qwen2.5-VL-7B (Latent)~\citep{bai2025qwen2_5_vl} & 83.5 & 29.9 & 12.6 \\
\midrule
\rowcolor{oursbg}\textbf{Ours (Gemma-4-4B)} & 96.7 & 72.1 & 64.5 \\
\rowcolor{oursbg}\textbf{Ours (Gemma-4-26B)} & \textbf{97.2} & \textbf{72.5} & \textbf{65.5} \\
\bottomrule
\end{tabularx}
\vspace{-3mm}
\end{table}

\noindent
{\bf Massive Object Counting on UNICBench \cite{unicbench2026}.}
\label{subsec:massive_counting}
Beyond single-target search, we evaluate object counting on UNICBench~\cite{unicbench2026}, which partitions scenes into \textit{Easy}, \textit{Medium}, and \textit{Hard} splits by target count and crowd density (\cref{tab:unicbench_results}). While foundation models perform well on the sparse \textit{Easy} split, reaching 74.4\%--76.8\% accuracy for Gemma-4-4B/26B~\cite{team2024gemma} and 85.5\% for Gemini 3.1 Pro~\cite{reid2024gemini}, their accuracy on the densely packed \textit{Hard} split drops to 9.9\%--10.1\% and 15.8\% as adjacent instances merge in the downsampled global view. Latent visual reasoning (Qwen2.5-VL-7B Latent~\cite{bai2025qwen2_5_vl}, 12.6\% on \textit{Hard}) cannot recover visual details already lost during encoding, while visual programming (ViperGPT~\cite{suris2023vipergpt}, 5.2\% on \textit{Hard}) suffers from compounding detector failures in crowds. 
By partitioning dense clusters into regional sub-crops and examining them independently, our method reaches \textbf{96.7\%} / \textbf{72.1\%} / \textbf{64.5\%} with Gemma-4-4B and \textbf{97.2\%} / \textbf{72.5\%} / \textbf{65.5\%} with Gemma-4-26B across \textit{Easy} / \textit{Medium} / \textit{Hard}, outperforming Gemini 3.1 Pro by \textbf{+49.7\%} (65.5\% vs.\ 15.8\%) and lifting its frozen 26B backbone by \textbf{+55.4\%} (from 10.1\%) on the \textit{Hard} split.

\noindent
{\bf Ultra-High-Resolution Scaling on HR-Bench \cite{wang2024hrbench}.}
\label{subsec:hrbench}
We further test how perception scales to 4K/8K resolutions on HR-Bench (\cref{tab:vstar_results}), which measures multiple-choice accuracy across Fine-grained Single-instance (FSP) and Cross-instance Perception (FCP). Under fixed visual token budgets, direct-reasoning VLMs degrade in overall accuracy from 4K to 8K, with GPT-4o~\cite{achiam2023gpt4} dropping from 59.0\% to 55.5\%, LLaVA-OneVision-7B~\cite{li2024llavaonevision} from 63.0\% to 59.8\%, and Qwen2.5-VL-32B~\cite{bai2025qwen2_5_vl} from 73.9\% to 70.4\%. Active visual search methods mitigate this drop, led by DeepEyes-7B~\cite{zheng2025deepeyes} (75.1\% / 72.6\%) and DeepEyesV2-7B~\cite{hong2025deepeyesv2} (77.9\% / 73.8\%) at 4K / 8K. Lifting its frozen Gemma-4-4B backbone (70.5\% / 67.2\%), our approach achieves \textbf{82.1\%} overall accuracy (\textbf{95.5\%} FSP) at 4K and \textbf{79.4\%} (\textbf{92.4\%} FSP) at 8K, surpassing DeepEyes-7B by \textbf{+7.0\%} and \textbf{+6.8\%} (and DeepEyesV2-7B by \textbf{+4.2\%} and \textbf{+5.6\%}).

\begin{table}[t]
  \centering
  \small
  \setlength{\tabcolsep}{4pt}
  \caption{\textbf{Visual Difference Reasoning} on VDiff-Bench-Gen~\citep{vdiffbench2026}. Baselines: official leaderboard. $^*$On our reproduction, as official data is unreleased; baselines match within 5\% on average.}
  \vspace{-2ex}
  \label{tab:vdiffbench}
  \begin{tabularx}{\linewidth}{Xcc}
    \toprule
    \textbf{Model} & \textbf{Micro-F1} & \textbf{Exact-set (\%)} \\
    \midrule
    \multicolumn{3}{l}{\textcolor{slateheader}{\textit{Open-Weight Vision-Language Models}}} \\
    LLaVA-OneVision-7B~\citep{li2024llavaonevision} & 7.1 & 1.3 \\
    InternVL3.5-8B & 12.0 & 5.7 \\
    Doubao Seed 1.6 Vision & 16.5 & 9.3 \\
    Qwen3-VL-8B Thinking & 17.3 & 2.7 \\
    Gemma-4-26B~\citep{team2024gemma} & 13.1$^*$ & 6.7$^*$ \\
    \midrule
    \multicolumn{3}{l}{\textcolor{slateheader}{\textit{Proprietary Frontier Multimodal Models}}} \\
    Grok 4.3 & 18.1 & 4.7 \\
    GPT-5.4 & 22.0 & 7.0 \\
    GPT-5.6 Terra & 24.7 & 7.7 \\
    Kimi K3 & 27.0 & 7.7 \\
    Gemini 3.1 Pro & 27.4 & 10.3 \\
    Gemini 2.5 Flash & 29.2 & 5.0 \\
    Gemini 3.5 Flash & \underline{46.8} & \underline{17.3} \\
    \midrule
    \rowcolor{oursbg}\textbf{Ours (Gemma-4-26B)} & \textbf{65.9}$^*$ & \textbf{24.0}$^*$ \\
    \bottomrule
  \end{tabularx}
  \vspace{-1ex}
\end{table}

\begin{figure*}[t]
    \centering
    \includegraphics[width=\textwidth]{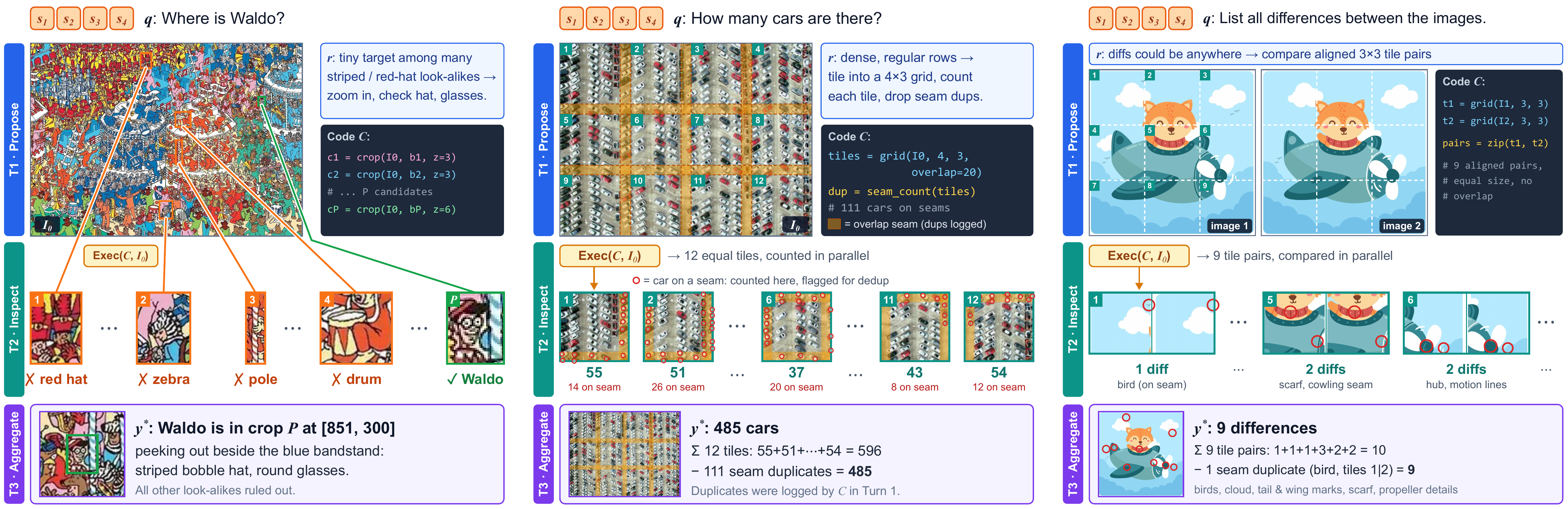}
    \vspace{-4ex}
    \caption{\textbf{Qualitative Examples Across Challenging Scenarios:} Visualization of three-turn search trajectories (\emph{Propose}, \emph{Inspect}, \emph{Aggregate}) generated by our framework on representative examples from WaldoBench (dense clutter search), UNICBench (massive counting), and VDiff-Bench (pairwise difference detection).}
    \label{fig:qualitative_visualizations}
    \vspace{-2ex}
\end{figure*}

\noindent
{\bf Discrepancy Detection on VDiff-Bench-Gen \cite{vdiffbench2026}.}
\label{subsec:vdiffbench}
Finally, we extend our evaluation to multi-image comparison (\cref{tab:vdiffbench}), which requires spotting subtle local differences across 300 high-resolution image pairs under viewpoint shifts and camera artifacts. Open-weight VLMs struggle to align fine details across images, with LLaVA-OneVision-7B~\cite{li2024llavaonevision} and Gemma-4-26B~\cite{team2024gemma} scoring only 7.1 and 13.1 Micro-F1 (1.3\% and 6.7\% Exact-set Accuracy), while Gemini 3.5 Flash leads proprietary baselines at 46.8 Micro-F1 (17.3\% Exact-set Accuracy). By cropping and comparing corresponding local regions across each image pair, our prefix-steered Gemma-4-26B achieves \textbf{65.9} Micro-F1 and \textbf{24.0\%} Exact-set Accuracy, surpassing the best-performing Gemini 3.5 Flash by \textbf{+19.1} Micro-F1 and its frozen backbone by \textbf{+52.8}.

\begin{figure}[t]
\centering
\resizebox{\linewidth}{!}{%
\sffamily
\begin{tikzpicture}
\definecolor{cVstar}{HTML}{1D4ED8}%
\definecolor{cVstarFill}{HTML}{3B82F6}
\definecolor{cHR4K}{HTML}{15803D}%
\definecolor{cHR4KFill}{HTML}{22C55E}
\definecolor{cHR8K}{HTML}{D97706}%
\definecolor{cHR8KFill}{HTML}{F59E0B}
\definecolor{cUNIC}{HTML}{B91C1C}%
\definecolor{cUNICFill}{HTML}{EF4444}

\definecolor{slatedark}{HTML}{334155}
\definecolor{gridgray}{HTML}{E2E8F0}
\definecolor{kfourbg}{HTML}{F1F5F9}
\definecolor{kfourline}{HTML}{94A3B8}

\node[draw=gridgray, fill=white, rounded corners=2pt, inner sep=2.8pt, font=\scriptsize, text=slatedark, align=center] at (3.92cm, 3.16cm) {
    \tikz[baseline=-0.5pt]{\draw[cVstar, line width=1.15pt] (0,0.06cm) -- (0.32cm,0.06cm); \draw[cVstar, fill=cVstarFill, line width=0.65pt] plot[mark=*, mark size=1.6pt] coordinates {(0.16cm,0.06cm)};}~$V^*$~Bench \hspace{6pt}
    \tikz[baseline=-0.5pt]{\draw[cHR4K, line width=1.15pt] (0,0.06cm) -- (0.32cm,0.06cm); \draw[cHR4K, fill=cHR4KFill, line width=0.65pt] plot[mark=square*, mark size=1.45pt] coordinates {(0.16cm,0.06cm)};}~HR-Bench 4K \hspace{6pt}
    \tikz[baseline=-0.5pt]{\draw[cHR8K, line width=1.15pt] (0,0.06cm) -- (0.32cm,0.06cm); \draw[cHR8K, fill=cHR8KFill, line width=0.65pt] plot[mark=triangle*, mark size=1.85pt] coordinates {(0.16cm,0.06cm)};}~HR-Bench 8K \hspace{6pt}
    \tikz[baseline=-0.5pt]{\draw[cUNIC, line width=1.15pt] (0,0.06cm) -- (0.32cm,0.06cm); \draw[cUNIC, fill=cUNICFill, line width=0.65pt] plot[mark=diamond*, mark size=1.95pt] coordinates {(0.16cm,0.06cm)};}~UNICBench {\tiny(\textit{Hard})}
};

\begin{axis}[
    name=leftplot,
    at={(0cm,0cm)},
    width=5.45cm,
    height=4.05cm,
    ymin=62.0, ymax=95.5,
    xmin=-0.48, xmax=5.42,
    xtick={0,1,2,3,4,5},
    xticklabels={$0$, $1$, $2$, $4$, $8$, $16$},
    ytick={65,75,85,95},
    xlabel={Prefix Tokens ($K$)},
    xlabel style={font=\scriptsize, yshift=1pt, text=slatedark},
    xticklabel style={font=\tiny, text=slatedark},
    yticklabel style={font=\tiny, text=slatedark},
    ymajorgrids=true,
    grid style={dashed, draw=gridgray, line width=0.5pt},
    axis lines*=left,
    tick align=outside,
    major tick length=1.8pt,
    axis line style={draw=slatedark, line width=0.65pt},
    tick style={draw=slatedark, line width=0.55pt},
    clip=false
]
\node[font=\tiny, text=slatedark, anchor=south west, inner sep=0pt] at (axis cs:-0.92,96.4) {Acc. (\%)};
\fill[kfourbg] (axis cs:2.70,62.45) rectangle (axis cs:3.30,95.0);
\draw[densely dashed, draw=kfourline, line width=0.65pt] (axis cs:3,62.7) -- (axis cs:3,74.6);
\draw[densely dashed, draw=kfourline, line width=0.65pt] (axis cs:3,86.0) -- (axis cs:3,92.0);

\addplot[color=cHR8K, line width=1.15pt, mark=triangle*, mark size=1.85pt, mark options={fill=cHR8KFill, draw=cHR8K, line width=0.6pt}] coordinates {
    (0,67.2) (1,74.9) (2,77.6) (3,79.4) (4,79.8) (5,80.1)
};
\node[font=\tiny, text=cHR8K, anchor=north, inner sep=1pt] at (axis cs:0,66.1) {67.2};
\node[font=\tiny\bfseries, text=cHR8K, anchor=north, inner sep=1pt] at (axis cs:3,78.3) {79.4};
\node[font=\tiny, text=cHR8K, anchor=north, inner sep=1pt] at (axis cs:5,79.0) {80.1};

\addplot[color=cHR4K, line width=1.15pt, mark=square*, mark size=1.45pt, mark options={fill=cHR4KFill, draw=cHR4K, line width=0.6pt}] coordinates {
    (0,70.5) (1,78.0) (2,80.4) (3,82.1) (4,82.5) (5,82.6)
};
\node[font=\tiny, text=cHR4K, anchor=south, inner sep=1pt] at (axis cs:-0.10,71.6) {70.5};
\node[font=\tiny\bfseries, text=cHR4K, anchor=south, inner sep=1pt] at (axis cs:3,82.8) {82.1};
\node[font=\tiny, text=cHR4K, anchor=south, inner sep=1pt] at (axis cs:5,83.2) {82.6};

\addplot[color=cVstar, line width=1.15pt, mark=*, mark size=1.6pt, mark options={fill=cVstarFill, draw=cVstar, line width=0.6pt}] coordinates {
    (0,76.6) (1,87.4) (2,90.1) (3,92.0) (4,92.3) (5,92.5)
};
\node[font=\tiny, text=cVstar, anchor=south, inner sep=1pt] at (axis cs:-0.10,77.9) {76.6};
\node[font=\tiny\bfseries, text=cVstar, anchor=south, inner sep=1pt] at (axis cs:3,92.6) {92.0};
\node[font=\tiny, text=cVstar, anchor=south, inner sep=1pt] at (axis cs:5,93.0) {92.5};
\end{axis}

\begin{axis}[
    name=rightplot,
    at={(4.32cm,0cm)},
    width=5.45cm,
    height=4.05cm,
    ymin=0, ymax=74,
    xmin=-0.48, xmax=5.42,
    xtick={0,1,2,3,4,5},
    xticklabels={$0$, $1$, $2$, $4$, $8$, $16$},
    ytick={0,20,40,60},
    xlabel={Prefix Tokens ($K$)},
    xlabel style={font=\scriptsize, yshift=1pt, text=slatedark},
    xticklabel style={font=\tiny, text=slatedark},
    yticklabel style={font=\tiny, text=slatedark},
    ymajorgrids=true,
    grid style={dashed, draw=gridgray, line width=0.5pt},
    axis lines*=left,
    tick align=outside,
    major tick length=1.8pt,
    axis line style={draw=slatedark, line width=0.65pt},
    tick style={draw=slatedark, line width=0.55pt},
    clip=false
]
\fill[kfourbg] (axis cs:2.70,0.95) rectangle (axis cs:3.30,72.8);
\draw[densely dashed, draw=kfourline, line width=0.65pt] (axis cs:3,1.5) -- (axis cs:3,65.5);

\addplot[color=cUNIC, line width=1.2pt, mark=diamond*, mark size=2.0pt, mark options={fill=cUNICFill, draw=cUNIC, line width=0.6pt}] coordinates {
    (0,10.1) (1,48.7) (2,58.9) (3,65.5) (4,66.1) (5,66.4)
};
\node[font=\tiny, text=cUNIC, anchor=west, inner sep=1pt] at (axis cs:0.16,10.1) {10.1};
\node[font=\tiny, text=cUNIC, anchor=south, inner sep=1pt] at (axis cs:0.92,50.5) {48.7};
\node[font=\tiny, text=cUNIC, anchor=south, inner sep=1pt] at (axis cs:1.95,60.5) {58.9};
\node[font=\tiny\bfseries, text=cUNIC, anchor=south, inner sep=1pt] at (axis cs:3,67.0) {65.5};
\node[font=\tiny, text=cUNIC, anchor=north, inner sep=1pt] at (axis cs:4,63.8) {66.1};
\node[font=\tiny, text=cUNIC, anchor=south, inner sep=1pt] at (axis cs:5,67.7) {66.4};

\end{axis}
\end{tikzpicture}%
}
\vspace{-4ex}
\caption{\small\textbf{Ablation on Prefix Token Count ($K$):} Effect of varying the number of learnable prefix tokens $\mathbf{S}$ across $V^*$~Bench, HR-Bench (4K / 8K), and UNICBench (\textit{Hard}). Accuracy saturates rapidly at our default $K=4$ (highlighted).}
\label{fig:ablation_k_tokens}
\vspace{-3mm}
\end{figure}
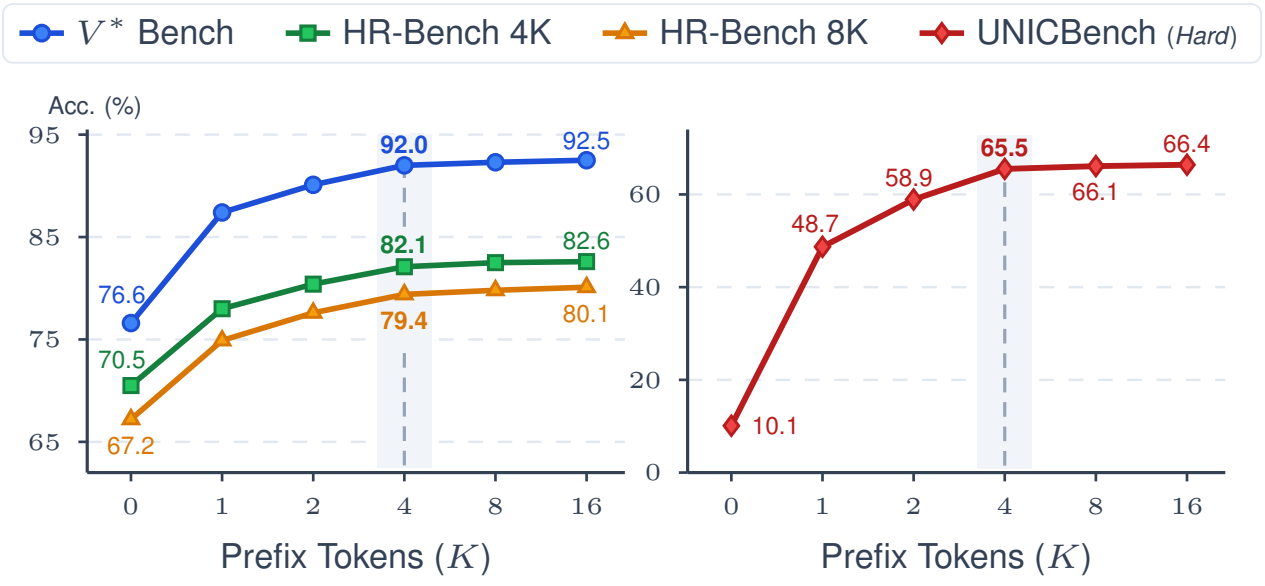

\noindent 
{\bf Qualitative Analysis.}
\cref{fig:qualitative_visualizations} illustrates how our three-turn trajectory adapts across tasks: Turn~1 proposes targeted candidate boxes on WaldoBench (left), a $4{\times}3$ grid on UNICBench (middle), and aligned $3{\times}3$ tile pairs on VDiff-Bench (right); Turn~2 inspects the crops in parallel to filter striped distractors, count local instances, or spot tile-level differences; and Turn~3 removes seam duplicates to produce the final prediction. 
Although each task is designed to address a distinct objective, such as localization, counting, or identifying differences, they share a common underlying structure that can be effectively accommodated within our framework. 
This shows our method's versatility and adaptability to support a broad range of tasks without requiring task-specific solutions.

\subsection{Ablation Studies and Analysis}
\label{subsec:ablation_studies}

\noindent
{\bf Number of Prefix Tokens.}
We vary the prefix length $K \in \{0, 1, 2, 4, 8, 16\}$ in~\cref{fig:ablation_k_tokens} to study how prefix capacity steers the frozen VLM. Compared to single-turn answering ($K{=}0$), even $K{=}1$ captures most of the initial gain (jumping from 76.6\% to 87.4\% on $V^*$~Bench, 70.5\%/67.2\% to 78.0\%/74.9\% on HR-Bench 4K/8K, and 10.1\% to 48.7\% on UNICBench \textit{Hard}), acting as a coarse trigger that switches the model from single-turn answering to our three-turn search. Since a single token imposes a low-rank bottleneck with only one steering direction in frozen self-attention, expanding the prefix to $K{=}4$ spans multiple directions to better coordinate the search policy, boosting accuracy to 92.0\%, 82.1\%/79.4\%, and 65.5\%.
Beyond $K{=}4$, accuracy increases only marginally ($\le 0.9\%$ at $K{=}8, 16$), as the remaining gap is bounded by the frozen VLM's perceptual limits (\eg, extremely small targets, camouflage, or severe occlusion) rather than prefix capacity.
Notably, if we instead force the frozen VLM to execute the three-turn workflow via text prompting alone without $\mathbf{S}$, accuracy collapses to 32.1\%, 35.4\%/30.0\%, and 3.8\% (well below single-turn answering), confirming that learned prefix tokens are essential to steer the search policy.

\begin{table}[t]
\centering
\small
\setlength{\tabcolsep}{4.0pt}
\caption{\textbf{Ablation on KV-Cache Context} (Gemma-4-4B).}
\label{tab:kv_ablation}
\vspace{-2ex}
\resizebox{0.95\columnwidth}{!}{%
\begin{tabular}{cccccc}
\toprule
\multicolumn{2}{c}{\textbf{Reused in Turn 2 \& 3}} & \multicolumn{2}{c}{\textbf{Reused in Turn 3}} & \textbf{$V^*$} & \textbf{UNIC} \\
\cmidrule(lr){1-2} \cmidrule(lr){3-4}
$(\mathbf{S}, \mathbf{V}_0)$ & Rationale $r$ & Code $\mathcal{C}$ & Patches $\mathbf{V}^{(p)}$ & \textbf{Bench} & \textbf{(\textit{Hard})} \\
\midrule
\textcolor{slateheader}{--} & \textcolor{slateheader}{--} & \textcolor{slateheader}{--} & \textcolor{slateheader}{--} & 59.2 & 40.3 \\
\checkmark & \textcolor{slateheader}{--} & \checkmark & \checkmark & 87.6 & 59.6 \\
\checkmark & \checkmark & \textcolor{slateheader}{--} & \checkmark & 91.5 & 60.8 \\
\checkmark & \checkmark & \checkmark & \textcolor{slateheader}{--} & 92.0 & 64.3 \\
\midrule
\rowcolor{oursbg}\checkmark & \checkmark & \checkmark & \checkmark & \textbf{92.0} & \textbf{64.5} \\
\bottomrule
\end{tabular}%
}
\vspace{-1ex}
\end{table}

\noindent
{\bf KV-Cache Context.}
\cref{tab:kv_ablation} ablates the components retained in the KV-cache across Turns~2 and~3. Without caching prior turns, accuracy collapses to 59.2\% on $V^*$~Bench and 40.3\% on UNICBench \textit{Hard}, as isolated crops lose both global spatial context $\mathbf{V}_0$ and prefix steering $\mathbf{S}$. Omitting the rationale $r$ also drops accuracy by 4.4\% and 4.9\%, showing that $r$ not only guides code generation in Turn~1, but also instructs downstream inspection and aggregation. In Turn~3, dropping $\mathcal{C}$ barely affects single-target search ($-0.5\%$) yet degrades dense counting by 3.7\%, where box coordinates are needed to reconcile counts across adjacent crops. Finally, removing the patches $\mathbf{V}^{(p)}$ leaves accuracy nearly unchanged (92.0\% and 64.3\%), indicating that Turn~2 already distills key visual details into text findings $\mathbf{a}^{(p)}$, though we retain $\mathbf{V}^{(p)}$ as it is already cached from Turn~2 at no extra cost.

\begin{table}[t]
\centering
\small
\setlength{\tabcolsep}{5.2pt}
\caption{\textbf{Latency and Token Efficiency} on HR-Bench 4K.}
\vspace{-2ex}
\label{tab:latency}
\resizebox{\linewidth}{!}{%
\begin{tabular}{lcccc}
\toprule
\textbf{Method} & \textbf{Turns} $\downarrow$ & \textbf{Vis.\ Tokens} $\downarrow$ & \textbf{Latency} $\downarrow$ & \textbf{Acc.\ (\%)} $\uparrow$ \\
\midrule
\textcolor{slateheader}{LLaVA-OV-7B~\cite{li2024llavaonevision}} & \textcolor{slateheader}{1.0} & \textcolor{slateheader}{8.1K} & \textcolor{slateheader}{0.8s} & \textcolor{slateheader}{63.0} \\
\midrule
ZoomEye~\cite{shen2024zoomeye} & 13.5 & 29.3K & 10.2s & 69.6 \\
DeepEyes-7B~\cite{zheng2025deepeyes} & 4.2 & 7.0K & 5.1s & 75.1 \\
\midrule
\rowcolor{oursbg}\textbf{Ours (Gemma-4-4B)} & \textbf{3.0} & \textbf{4.6K} & \textbf{2.8s} & \textbf{82.1} \\
\bottomrule
\end{tabular}%
}
\vspace{-1ex}
\end{table}

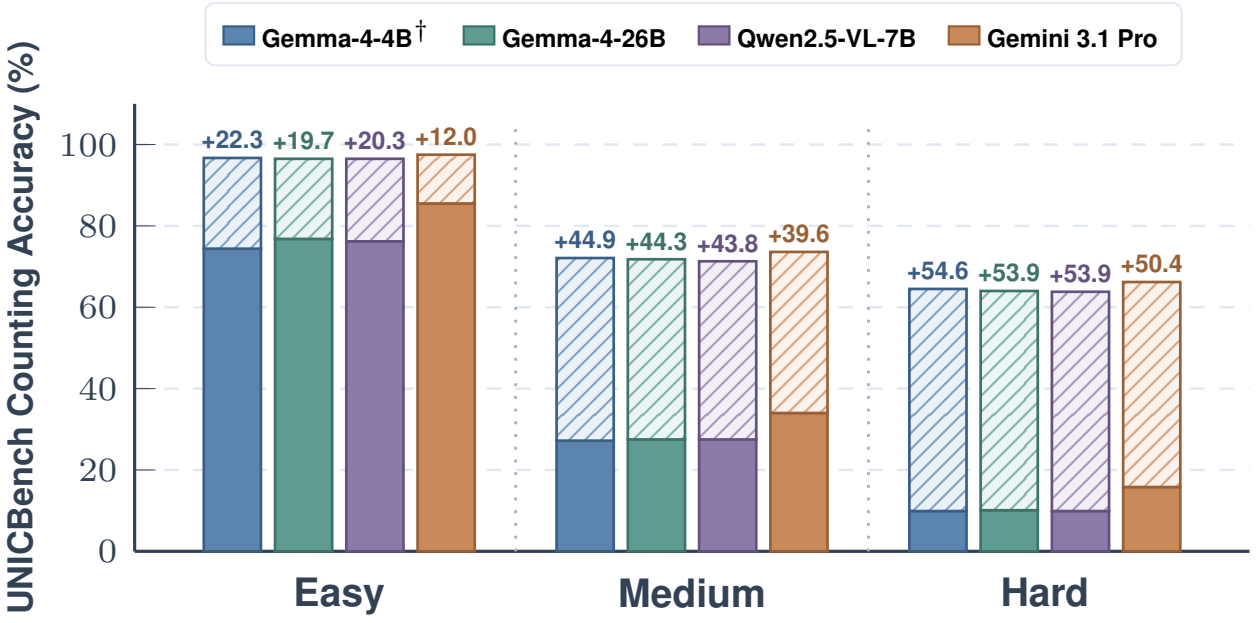
\begin{figure}[t]
\centering
\resizebox{\linewidth}{!}{%
\sffamily
\begin{tikzpicture}
\definecolor{mOneBase}{HTML}{5B84B1}%
\definecolor{mOneTop}{HTML}{EBF2F7}
\definecolor{mOneLine}{HTML}{3B5E85}

\definecolor{mTwoBase}{HTML}{5E9C91}%
\definecolor{mTwoTop}{HTML}{EBF5F3}
\definecolor{mTwoLine}{HTML}{3D7369}

\definecolor{mThreeBase}{HTML}{8C7AA9}%
\definecolor{mThreeTop}{HTML}{F2EFF6}
\definecolor{mThreeLine}{HTML}{63527F}

\definecolor{mFourBase}{HTML}{C98A5B}%
\definecolor{mFourTop}{HTML}{FBF2EB}
\definecolor{mFourLine}{HTML}{996035}

\definecolor{slatedark}{HTML}{334155}
\definecolor{gridgray}{HTML}{E2E8F0}

\newcommand{\DrawSingleBar}[6]{%
    \filldraw[fill=#6Base, draw=#6Line, line width=0.55pt] (axis cs:#1,0) rectangle (axis cs:#2,#4);
    \fill[fill=#6Top] (axis cs:#1,#4) rectangle (axis cs:#2,#5);
    \fill[pattern=north east lines, pattern color=#6Base!75] (axis cs:#1,#4) rectangle (axis cs:#2,#5);
    \draw[draw=#6Line, line width=0.55pt] (axis cs:#1,#4) rectangle (axis cs:#2,#5);
    \node[font=\tiny\bfseries, text=#6Line, anchor=south, inner sep=1pt] at (axis cs:#3,#5+0.8) {\pgfmathparse{#5-#4}+\pgfmathprintnumber[assume math mode=true, fixed, fixed zerofill, precision=1]{\pgfmathresult}};
}
\newcommand{\PlotBarOne}[6]{%
    \DrawSingleBar{0.616}{0.778}{0.697}{#1}{#2}{mOne}%
    \DrawSingleBar{1.616}{1.778}{1.697}{#3}{#4}{mOne}%
    \DrawSingleBar{2.616}{2.778}{2.697}{#5}{#6}{mOne}%
}
\newcommand{\PlotBarTwo}[6]{%
    \DrawSingleBar{0.818}{0.980}{0.899}{#1}{#2}{mTwo}%
    \DrawSingleBar{1.818}{1.980}{1.899}{#3}{#4}{mTwo}%
    \DrawSingleBar{2.818}{2.980}{2.899}{#5}{#6}{mTwo}%
}
\newcommand{\PlotBarThree}[6]{%
    \DrawSingleBar{1.020}{1.182}{1.101}{#1}{#2}{mThree}%
    \DrawSingleBar{2.020}{2.182}{2.101}{#3}{#4}{mThree}%
    \DrawSingleBar{3.020}{3.182}{3.101}{#5}{#6}{mThree}%
}
\newcommand{\PlotBarFour}[6]{%
    \DrawSingleBar{1.222}{1.384}{1.303}{#1}{#2}{mFour}%
    \DrawSingleBar{2.222}{2.384}{2.303}{#3}{#4}{mFour}%
    \DrawSingleBar{3.222}{3.384}{3.303}{#5}{#6}{mFour}%
}

\begin{axis}[
    width=9.6cm,
    height=4.8cm,
    ymin=0, ymax=110,
    xmin=0.42, xmax=3.58,
    xtick={1,2,3},
    xticklabels={\textbf{Easy}, \textbf{Medium}, \textbf{Hard}},
    ytick={0,20,40,60,80,100},
    ylabel={\textbf{UNICBench Counting Accuracy (\%)}},
    ylabel style={font=\scriptsize, yshift=-3pt, text=slatedark},
    xticklabel style={font=\footnotesize, yshift=-2pt, text=slatedark},
    yticklabel style={font=\scriptsize, text=slatedark},
    ymajorgrids=true,
    grid style={dashed, draw=gridgray, line width=0.5pt},
    axis lines*=left,
    axis line style={draw=slatedark, line width=0.65pt},
    tick style={draw=slatedark},
    xtick style={draw=none},
    clip=false
]

\draw[dotted, draw=slatedark!40, line width=0.6pt] (axis cs:1.5,0) -- (axis cs:1.5,105);
\draw[dotted, draw=slatedark!40, line width=0.6pt] (axis cs:2.5,0) -- (axis cs:2.5,105);

\PlotBarOne  {74.4}{96.7} {27.2}{72.1} {9.9}{64.5}%
\PlotBarTwo  {76.8}{96.5} {27.5}{71.8} {10.1}{64.0}%
\PlotBarThree{76.2}{96.5} {27.5}{71.3} {9.9}{63.8}%
\PlotBarFour {85.5}{97.5} {34.0}{73.6} {15.8}{66.2}%

\end{axis}

\node[draw=gridgray, fill=white, rounded corners=2pt, inner sep=3.5pt, font=\tiny, align=center] at (4.02cm, 3.72cm) {
    \tikz[sharp corners, baseline=-0.5pt]{\filldraw[fill=mOneBase, draw=mOneLine, line width=0.45pt] (0,0) rectangle (0.22cm,0.13cm);}~\textbf{Gemma-4-4B$^\dagger$} \quad
    \tikz[sharp corners, baseline=-0.5pt]{\filldraw[fill=mTwoBase, draw=mTwoLine, line width=0.45pt] (0,0) rectangle (0.22cm,0.13cm);}~\textbf{Gemma-4-26B} \quad
    \tikz[sharp corners, baseline=-0.5pt]{\filldraw[fill=mThreeBase, draw=mThreeLine, line width=0.45pt] (0,0) rectangle (0.22cm,0.13cm);}~\textbf{Qwen2.5-VL-7B} \quad
    \tikz[sharp corners, baseline=-0.5pt]{\filldraw[fill=mFourBase, draw=mFourLine, line width=0.45pt] (0,0) rectangle (0.22cm,0.13cm);}~\textbf{Gemini 3.1 Pro}
};
\end{tikzpicture}%
}
\vspace{-4ex}
\caption{\textbf{Plug-in Transferability Study on UNICBench:} Zero-shot augmentation of diverse downstream VLMs across \textit{Easy}, \textit{Medium}, and \textit{Hard} counting splits ($^\dagger$source model on which prefix tokens are trained). Solid base bars denote direct VLM performance without our front-end, while hatched top segments show the accuracy gain ($+\Delta\%$) achieved when supplying dynamic Turn-1 crops from our prefix-steered Gemma-4-4B.}
\label{fig:plugin_transfer}
\end{figure}

\noindent
{\bf Inference Latency and Token Efficiency.}
In~\cref{tab:latency}, we compare end-to-end efficiency on HR-Bench 4K against single-turn native tiling and sequential visual search. Compared to LLaVA-OneVision-7B~\cite{li2024llavaonevision}, selective cropping inspects only task-relevant regions, processing 43\% fewer visual tokens (4.6K vs.\ 8.1K) while improving accuracy by +19.1\% (82.1\% vs.\ 63.0\%). Compared to open-ended sequential search in ZoomEye~\cite{shen2024zoomeye} (13.5 turns, 10.2s) and DeepEyes-7B~\cite{zheng2025deepeyes} (4.2 turns, 5.1s), our deterministic three-turn workflow reduces latency by \textbf{3.6$\times$} and \textbf{1.8$\times$} (to \textbf{2.8s}) and uses 34\% fewer visual tokens than DeepEyes-7B while achieving +7.0\% higher accuracy.

\noindent 
{\bf Cross-Model Transferability.}
While the learned prefix tokens $\mathbf{S}$ are tied to a specific backbone and cannot be directly transferred across models, the search policy $(r, \mathcal{C})$ externalized in Turn~1 is model-agnostic. In~\cref{fig:plugin_transfer}, we use our prefix-steered Gemma-4-4B as a Turn~1 front-end and feed its outputs directly to unmodified downstream VLMs (Gemma-4-4B, Gemma-4-26B~\cite{team2024gemma}, Qwen2.5-VL-7B~\cite{bai2025qwen2_5_vl}, and Gemini~3.1~Pro~\cite{reid2024gemini}) for zero-shot inspection and aggregation on UNICBench~\cite{unicbench2026}. Across all four models, our 4B Turn~1 policy transfers seamlessly, yielding average gains of +18.6\% on \textit{Easy}, \textbf{+43.2\%} on \textit{Medium}, and \textbf{+53.2\%} on \textit{Hard}. Notably, once guided by the same Turn~1 policy, all four models, from 4B to frontier Gemini~3.1~Pro, converge to nearly identical accuracy within every tier ($\approx 96.8\%$, $72.2\%$, and $64.6\%$), demonstrating that the primary bottleneck is neither model architecture nor parameter scale, but finding an effective search policy.

\section{Conclusion}
\label{sec:conclusion}
\vspace{-0.5ex}

We presented a framework for fine-grained visual reasoning that collapses open-ended, sequential visual search into a deterministic three-turn trajectory: proposing candidate crops upfront (\emph{Propose}), inspecting high-resolution patches in parallel (\emph{Inspect}), and aggregating local observations into a final answer (\emph{Aggregate}).
Our formulation eliminates iterative trial-and-error while substantially reducing inference latency. 
Leveraging capabilities of pre-trained VLMs, we steer the entire trajectory on a completely frozen backbone by optimizing only four learnable prefix tokens ($\approx 10.2\text{K}$ parameters, $20.5\text{ KB}$). 
Across five fine-grained perception benchmarks covering cluttered visual search, small-object grounding, dense counting, 4K/8K scaling, and multi-image difference detection, our method enables compact open-weight models to outperform both multi-turn RL methods and proprietary frontier models, while functioning as a zero-shot, plug-and-play visual front-end for off-the-shelf foundation models.

{
    \small
    \bibliographystyle{ieeenat_fullname}
    \bibliography{main}
}

\clearpage
\maketitlesupplementary

\appendix
\renewcommand{\thesection}{\Alph{section}}
\renewcommand{\thetable}{S\arabic{table}}
\renewcommand{\thefigure}{S\arabic{figure}}
\renewcommand{\theequation}{S\arabic{equation}}
\setcounter{table}{0}
\setcounter{figure}{0}
\setcounter{equation}{0}
\renewcommand{\topfraction}{0.95}
\renewcommand{\dbltopfraction}{0.95}
\renewcommand{\textfraction}{0.04}

\providecommand{\Paragraph}[1]{\vspace{1mm}\noindent\textbf{#1}}
\definecolor{slateheader}{HTML}{64748B}
\definecolor{promptbg}{HTML}{F8FAFC}
\definecolor{promptframe}{HTML}{CBD5E1}
\definecolor{prompttitle}{HTML}{1E293B}
\definecolor{pillSbg}{HTML}{DBEAFE}
\definecolor{pillSbd}{HTML}{60A5FA}
\definecolor{pillUbg}{HTML}{FEF3C7}
\definecolor{pillUbd}{HTML}{F59E0B}
\definecolor{pillKbg}{HTML}{E2E8F0}
\definecolor{pillKbd}{HTML}{94A3B8}
\definecolor{pillObg}{HTML}{DCFCE7}
\definecolor{pillObd}{HTML}{4ADE80}
\tcbset{coltitle=prompttitle, colbacktitle=promptframe!55!white}
\newcommand{\tokpill}[3]{\tcbox[on line, arc=1.5pt, colback=#1, colframe=#2, boxrule=0.45pt, boxsep=0pt, left=3pt, right=3pt, top=1.5pt, bottom=1.5pt]{\scriptsize\strut #3}}
\newcommand{\pillS}[1]{\tokpill{pillSbg}{pillSbd}{#1}}
\newcommand{\pillUser}[1]{\tokpill{pillUbg}{pillUbd}{#1}}
\newcommand{\pillKV}[1]{\tokpill{pillKbg}{pillKbd}{#1}}
\newcommand{\pillRun}[1]{\tokpill{white}{pillKbd}{#1}}
\newcommand{\pillOut}[1]{\tokpill{pillObg}{pillObd}{#1}}

\noindent
This supplementary material provides training configurations, data curation and baseline comparisons, turn-level token structure and inference workflow, trajectory visualizations, and failure analysis:
\begin{itemize}
    \item \textbf{\cref{supp:sec_setup}:} Training hyperparameters and prefix parameterization (\cref{tab:supp_hyperparams}).
    \item \textbf{\cref{supp:sec_curation}:} Data curation pipeline, dataset statistics, and example trajectories (\cref{supp:subsec_curation,tab:supp_dataset_stats,fig:supp_data_example,fig:supp_exec_code}), together with a comparison of training methods, data scale, and trajectory annotations against baseline models (\cref{supp:subsec_baseline_comp,tab:supp_baseline_data}).
    \item \textbf{\cref{supp:sec_prompts}:} Turn-by-turn input token structure, KV-cache reuse, and explicit system prompt for the text-prompting baseline (\cref{fig:supp_prompts,fig:supp_baseline_prompt}).
    \item \textbf{\cref{supp:sec_visualizations}:} Three-turn trajectory visualizations across five scenarios (\cref{fig:supp_traj_waldo,fig:supp_traj_counting,fig:supp_traj_multiclue,fig:supp_traj_hrbench,fig:supp_traj_vdiff}), covering dense clutter, object counting, relational disambiguation, 8K perception, and multi-image diffing.
    \item \textbf{\cref{supp:sec_failures}:} Qualitative failure cases (\cref{fig:supp_failure_cases}), limitations, future work, and broader impact.
\end{itemize}

\section{Training Configurations}
\label{supp:sec_setup}

\cref{tab:supp_hyperparams} summarizes the prefix parameterization and optimization hyperparameters across different VLM backbones. For each backbone, the vision encoder, multimodal projector, and LLM decoder remain completely frozen, and only the $K=4$ continuous prefix tokens $\mathbf{S} \in \mathbb{R}^{K \times D}$ (20.5--28.7\,KB in BFloat16) are updated under the same optimization schedule.

\begin{table}[!b]
\centering
\footnotesize
\renewcommand{\arraystretch}{0.98}
\caption{\textbf{Training Configuration and Hyperparameters} across different VLM backbones.}
\vspace{-1ex}
\label{tab:supp_hyperparams}
\begin{tabular*}{\linewidth}{@{\extracolsep{\fill}} l l @{}}
\toprule
\textbf{Setting} & \textbf{Specification} \\
\midrule
\multicolumn{2}{@{}l}{\textcolor{slateheader}{\textit{Prefix Parameterization ($K=4$, BFloat16)}}} \\
Gemma-4-4B~\cite{team2024gemma} & $D{=}2560$, 10.2K params (20.5\,KB) \\
Qwen2.5-VL-7B~\cite{bai2025qwen2_5_vl} & $D{=}3584$, 14.3K params (28.7\,KB) \\
Gemma-4-26B~\cite{team2024gemma} & $D{=}2816$, 11.3K params (22.5\,KB) \\
Prefix Init.\ ($\mathbf{S}_{0}$) & $\mathcal{N}(0, 1)$ (PyTorch default) \\
Frozen Modules & Vision encoder, projector, LLM (100\%) \\
\midrule
\multicolumn{2}{@{}l}{\textcolor{slateheader}{\textit{Optimization}}} \\
Optimizer & AdamW ($\beta_1{=}0.9, \beta_2{=}0.99$) \\
Learning Rate & $2{\times}10^{-5} \to 1{\times}10^{-6}$ (cosine) \\
Weight Decay / Clip & 0.05 / max grad norm 1.0 \\
Warmup / Epochs & 5\% linear / 3 epochs \\
Batch Size & 64 trajectories (8 per GPU) \\
Compute & 8 $\times$ NVIDIA A100 (FSDP) \\
\bottomrule
\end{tabular*}
\vspace{-1ex}
\end{table}

\begin{table}[!t]
\centering
\scriptsize
\setlength{\tabcolsep}{1.7pt}
\renewcommand{\arraystretch}{1.12}
\caption{\textbf{Summary of the Curated Training Dataset ($\mathcal{D}$).} For each benchmark, we list the source of the training images, the number of human-annotated trajectory seeds, the total number of training samples, and the average number of crops $P$ per trajectory. $^\dagger$Averaged across WaldoBench and $V^*$~Bench.}
\vspace{-1ex}
\label{tab:supp_dataset_stats}
\begin{tabularx}{\linewidth}{l >{\raggedright\arraybackslash}X c c c}
\toprule
\textbf{Target} & \textbf{Source} & \textbf{Trajectory} & \textbf{Total} & \textbf{Avg.} \\[-0.3ex]
\textbf{Benchmark} & \textbf{Images} & \textbf{Seeds} & \textbf{Samples} & \textbf{Crops ($P$)} \\
\midrule
\multirow{2}{*}{WaldoBench~\cite{waldobench2026}} & \mbox{Roboflow Waldo~\cite{roboflow_waldo}} & \multirow{2}{*}{$20$} & \multirow{2}{*}{$200$} & \multirow{3}{*}{$8.9^\dagger$} \\[-0.3ex]
 & + synthesis & & & \\
$V^*$~Bench~\cite{wu2024visualsearch} & COCO-2017~\cite{lin2014coco} & $30$ & $300$ & \\
HR-Bench~\cite{wang2024hrbench} & DIV8K~\cite{gu2019div8k} & $30$ & $300$ & $4.6$ \\
UNICBench~\cite{unicbench2026} & FSC-147~\cite{ranjan2021learning} & $30$ & $300$ & $11.4$ \\
VDiff-Bench-Gen~\cite{vdiffbench2026} & Synthesis & $30$ & $300$ & $12.2$ \\
\midrule
\rowcolor{oursbg}\textbf{Total ($\mathcal{D}$)} & -- & \textbf{140} & \textbf{1,400} & \textbf{9.2} \\
\bottomrule
\end{tabularx}
\vspace{-1ex}
\end{table}

\section{Data Curation and Baseline Comparison}
\label{supp:sec_curation}

\subsection{Data Curation Pipeline}
\label{supp:subsec_curation}

To tune the prefix tokens $\mathbf{S}$ for our proposed visual search, we curate a unified dataset $\mathcal{D}$ with three-turn trajectory annotations across five different tasks in three stages:
\begin{itemize}
    \item \textbf{Stage 1 (Source Collection):} For each task, we collect training images that do not overlap with the benchmark's evaluation set. Where public images are limited or unavailable for certain tasks, we use an LLM to synthesize additional images and manually verify that the synthesized images have no data leakage from the evaluation samples.
    \item \textbf{Stage 2 (Human Seed Annotation):} We manually annotate a small set of seed images with complete three-turn trajectories (\emph{Trajectory Seeds}). Specifically, each trajectory $\tau = (r, \mathcal{C}, \{\mathbf{a}^{(p)}\}_{p=1}^P, \mathbf{y}^*)$ comprises the Turn~1 rationale $r$ and Python crop code $\mathcal{C}$ (crop boxes and zoom-in factors), the Turn~2 regional findings $\{\mathbf{a}^{(p)}\}_{p=1}^P$, and the Turn~3 final answer $\mathbf{y}^*$. \cref{fig:supp_data_example,fig:supp_exec_code} show a complete training example and Python cropping script.
    \item \textbf{Stage 3 (LLM Trajectory Scaling):} Using the human-annotated trajectory seeds as in-context demonstrations, the LLM generates three-turn trajectories in the same format for the remaining images.
\end{itemize}

\begin{figure*}[!t]
\centering
\begin{tcolorbox}[colback=promptbg, colframe=promptframe, boxrule=0.6pt, arc=2pt, left=6pt, right=6pt, top=5pt, bottom=5pt, title={\textbf{\footnotesize Example Curated Training Trajectory $\tau = (r, \mathcal{C}, \{\mathbf{a}^{(p)}\}_{p=1}^P, \mathbf{y}^*) \in \mathcal{D}$}}]
\scriptsize
\begin{verbatim}
{
  "sample_id": "waldo_train_0137",
  "image": "waldo/train/0137.jpg",
  "query": "Where is Waldo? Return his location as a point.",
  "turn_1_propose": {
    "rationale": "The scene is a crowded beach with hundreds of small figures. Waldo wears a red-and-white
                  striped shirt, a bobble hat, and round glasses. Red-and-white patterns appear near the
                  ice-cream stand (upper left), the row of deck chairs (center), and the pier (lower right),
                  so I zoom into these three regions. In Turn 2, check each crop for the full outfit and
                  reject look-alikes such as striped umbrellas and towels. In Turn 3, keep the crop that
                  matches and map its location back to the full image.",
    "code": "crops = [crop(I_0, box=[92, 141, 236, 289], z=4.0),
                      crop(I_0, box=[418, 402, 566, 548], z=4.0),
                      crop(I_0, box=[701, 655, 842, 803], z=4.0)]"
  },
  "turn_2_inspect_parallel": [
    {"crop_id": 1, "box": [92, 141, 236, 289],
     "finding": "Not found. The red-and-white stripes belong to the awning of the ice-cream stand."},
    {"crop_id": 2, "box": [418, 402, 566, 548],
     "finding": "Found. A figure behind the second deck chair wears a red-and-white striped shirt, a bobble hat,
                 and round glasses, at about [0.56, 0.41] within this crop."},
    {"crop_id": 3, "box": [701, 655, 842, 803],
     "finding": "Not found. A man in a striped shirt stands on the pier, but he has no hat or glasses."}
  ],
  "turn_3_aggregate": "Only crop 2 contains a figure with the full outfit; crops 1 and 3 are distractors.
                       Mapping the crop location back to the full image: <answer>[501, 462]</answer>."
}
\end{verbatim}
\end{tcolorbox}
\caption{\textbf{Example Training Trajectory from $\mathcal{D}$} (WaldoBench). Each sample stores the query, the Turn~1 rationale $r$ and crop code $\mathcal{C}$ (helper definitions omitted for brevity; see \cref{fig:supp_exec_code}), the per-crop Turn~2 findings $\mathbf{a}^{(p)}$, and the Turn~3 final answer $\mathbf{y}^*$; during SFT, each JSON record is formatted into the three-turn sequence in \cref{fig:supp_prompts}. Long strings are wrapped for display.}
\label{fig:supp_data_example}
\vspace{-1ex}
\end{figure*}

We use Gemini 3.1 Pro~\cite{reid2024gemini} as the LLM for all five benchmarks and combine all 1,400 samples into a single dataset $\mathcal{D}$ to train one prefix $\mathbf{S}$ per VLM backbone (\cref{tab:supp_dataset_stats}), as detailed below.

\vspace{0.5ex}\noindent
{\bf WaldoBench.}
For training on WaldoBench~\cite{waldobench2026}, we use 20 real \emph{Where's Waldo} images from the public Roboflow Universe dataset~\cite{roboflow_waldo} together with 180 synthesized \emph{Where's Waldo} images (200 samples in total), and verify generated trajectories by checking that the predicted point in normalized image coordinates falls within 5\% of the ground-truth Waldo position. For evaluation, we follow the WaldoBench testing pipeline on the 53 test images under the same 5\% hit rule, where image enhancement is applied only to remove existing bounding-box overlays and improve image quality via super-resolution without changing the scene content.

\vspace{0.5ex}\noindent
{\bf $V^*$~Bench.}
Since $V^*$~Bench~\cite{wu2024visualsearch} provides only an evaluation set, we follow SEAL~\cite{wu2024visualsearch} and sample 300 training images from COCO-2017~\cite{lin2014coco}, choosing the 30 seeds to cover diverse object categories. Focusing on fine-grained attributes and spatial relations of small objects, these samples together with WaldoBench average $P = 8.9$ candidate crops per trajectory.

\begin{figure}[!t]
\centering
\begin{tcolorbox}[colback=promptbg, colframe=promptframe, boxrule=0.6pt, arc=2pt, left=5pt, right=5pt, top=3pt, bottom=3pt, title={\textbf{\footnotesize Example Python Cropping Script ($\mathcal{C}$) Generated in Turn 1}}]
\linespread{0.92}\scriptsize
\begin{verbatim}
from PIL import Image

def crop(img, box, z=1.0):
    W, H = img.size
    x1, y1, x2, y2 = [min(max(v, 0), 1000)
                      for v in box]
    x1, x2 = sorted((x1, x2))
    y1, y2 = sorted((y1, y2))
    l, r = round(x1 * W / 1e3), round(x2 * W / 1e3)
    t, b = round(y1 * H / 1e3), round(y2 * H / 1e3)
    r, b = max(r, l + 1), max(b, t + 1)
    patch = img.crop((l, t, r, b))
    if z != 1.0:
        w, h = patch.size
        w, h = max(1,round(w*z)), max(1,round(h*z))
        patch = patch.resize((w, h), Image.BICUBIC)
    return patch

def grid(img, rows, cols, z=1.0):
    dx, dy = 1000 / cols, 1000 / rows
    boxes = [[j*dx, i*dy, (j+1)*dx, (i+1)*dy]
        for i in range(rows) for j in range(cols)]
    return [crop(img, box, z) for box in boxes]

crops = [crop(I_0, box=[412, 230, 508, 334], z=4.0),
         crop(I_0, box=[655, 120, 760, 248], z=3.0)]
\end{verbatim}
\end{tcolorbox}
\vspace{-1.5mm}
\caption{\textbf{Example Python Cropping Script ($\mathcal{C}$) Generated in Turn~1.} At each inference run, the VLM generates an executable script $\mathcal{C}$ tailored to the query (\eg, box-based, point-centered, or grid-based cropping on native $I_0$ with per-region zoom $z_p$), and $\mathrm{Exec}(\mathcal{C}, I_0)$ returns the patches $\{I^{(p)}\}_{p=1}^P$ for Turn~2.}
\label{fig:supp_exec_code}
\vspace{-1.5ex}
\end{figure}

\begin{table*}[!t]
\centering
\footnotesize
\setlength{\tabcolsep}{7.0pt}
\renewcommand{\arraystretch}{1.08}
\caption{\textbf{Training Data Comparison} across baselines in \cref{tab:hrbench_4k_results,tab:unicbench_results}. \textit{\# Samples / \# Trajectories} reports the total number of training samples and the number of search trajectory annotations for SFT; \textit{Scene Search Data} lists the training datasets used for scene search tasks (evaluated on $V^*$~Bench and HR-Bench); and \textit{Trajectory Source} indicates how each method acquires trajectory annotations ($^\dagger$with human-annotated seeds). Our method achieves stronger accuracy using a small fraction of the training data and annotations, highlighting the data efficiency of deterministic three-turn search.}
\label{tab:supp_baseline_data}
\begin{tabularx}{\textwidth}{l c c >{\raggedright\arraybackslash}X c}
\toprule
\textbf{Model} & \textbf{Training Method} & \textbf{\# Samples / \# Trajectories} & \textbf{Scene Search Data} & \textbf{Trajectory Source} \\[-0.3ex]
& & & \textcolor{slateheader}{\scriptsize ($V^*$~Bench, HR-Bench)} & \\
\midrule
\multicolumn{5}{l}{\textcolor{slateheader}{\textit{Latent Visual Reasoning Baselines}}} \\
Qwen2.5-VL-7B (Latent)~\cite{bai2025qwen2_5_vl,li2026lvr} & SFT + RL & 477K / 438K & GQA, Flickr30K, OpenImages & GT boxes \\
\midrule
\multicolumn{5}{l}{\textcolor{slateheader}{\textit{Active Visual Search Methods}}} \\
SEAL~\cite{wu2024visualsearch} & SFT & 387K / 267K & $V^*$-train (COCO, GQA, VAW) & LLM \\
Pixel-Reasoner (7B)~\cite{su2025pixelreasoner} & SFT + RL & 23K / 8.4K & Filtered $V^*$-train, SA-1B & LLM \\
DeepEyes (7B)~\cite{zheng2025deepeyes} & RL & 47K / 0 & Filtered $V^*$-train & N/A \\
DeepEyesV2 (7B)~\cite{hong2025deepeyesv2} & SFT + RL & 147K / 65K & Filtered $V^*$-train, SA-1B & LLM \\
CodeV (7B)~\cite{hou2026codev} & SFT + RL & 391K / 253K & Filtered $V^*$-train, Thyme-SFT & LLM \\
\midrule
\rowcolor{oursbg}\textbf{Ours} & SFT & \textbf{1.4K / 1.4K} & COCO, DIV8K & LLM$^\dagger$ \\
\bottomrule
\end{tabularx}
\end{table*}

\vspace{0.5ex}\noindent
{\bf HR-Bench.}
HR-Bench~\cite{wang2024hrbench} likewise provides only evaluation sets (4K and 8K splits), constructed from high-resolution images in DIV8K~\cite{gu2019div8k} and the Internet; for training, we sample 300 images from the remaining DIV8K images outside HR-Bench 4K/8K. Since its queries typically target one or two localized instances anchored by clear global landmarks rather than crowded distractors or dense grids, trajectories require fewer regions ($P = 4.6$ crops on average).

\vspace{0.5ex}\noindent
{\bf UNICBench.}
For dense object counting on UNICBench~\cite{unicbench2026}, we sample 300 training images from FSC-147~\cite{ranjan2021learning} and verify generated trajectories by requiring the predicted count to fall within 5\% of the ground-truth count. Partitioning crowded scenes into sub-regions or regular grids yields $P = 11.4$ crops on average.

\vspace{0.5ex}\noindent
{\bf VDiff-Bench-Gen.}
As the official image pairs of VDiff-Bench-Gen~\cite{vdiffbench2026} are not publicly available, we follow its difference-synthesis pipeline to generate 300 training image pairs (averaging $P = 12.2$ aligned crop pairs per trajectory) and 300 test image pairs matching the official test size. To verify that our reproduced test set has similar difficulty to the official benchmark, we evaluate leaderboard baselines on our 300 test pairs and obtain scores within 5\% on average of their reported numbers (models evaluated on our test set are marked with $^*$ in \cref{tab:vdiffbench}).

\subsection{Comparison with Baseline Methods}
\label{supp:subsec_baseline_comp}

\cref{tab:supp_baseline_data} compares all trained baselines in \cref{tab:hrbench_4k_results,tab:unicbench_results} (excluding off-the-shelf VLMs and training-free methods), where each method trains one shared model across all benchmarks. Previous methods update full backbones over \textbf{23K--477K} samples, whereas our proposed method requires only \textbf{1.4K} SFT samples ($16\times$--$341\times$ fewer) while achieving higher accuracy. While the \textit{Scene Search Data} column in \cref{tab:supp_baseline_data} lists the datasets used for scene search, the total sample counts of full-backbone methods (\eg, SEAL~\cite{wu2024visualsearch}, DeepEyes~\cite{zheng2025deepeyes}) also include large general datasets such as LLaVA-Instruct and ArXivQA to prevent forgetting general VLM capabilities. Fine-tuning prefix tokens on a frozen VLM avoids this issue altogether, using 1.4K samples in total and only 0.6K scene-search samples (300 COCO + 300 DIV8K for $V^*$~Bench and HR-Bench; \cref{tab:supp_dataset_stats}).

\begin{figure*}[!t]
\centering
\begin{tcolorbox}[colback=promptbg, colframe=promptframe, boxrule=0.6pt, arc=2pt, left=6pt, right=6pt, top=4pt, bottom=4pt, title={\textbf{\footnotesize Example of Three-Turn Inference Token Sequences ($P=2$ Crops)}}]
\footnotesize\frenchspacing
{\scriptsize \textbf{Token Blocks:}\;\;
\pillS{Learned Prefix $\mathbf{S}$} \;\;
\pillUser{User Input (Turn 1 only)} \;\;
\pillKV{Reused KV-Cache} \;\;
\pillRun{Turn Prompt} \;\;
\pillOut{VLM Decoded Output}}
\vspace{3pt}\hrule\vspace{3.5pt}
\textbf{1. Turn 1 (\emph{Propose}):} \quad {\scriptsize\textcolor{slateheader}{User provides $(I_0, \mathbf{q})$ once; input + decoded output are cached as $\text{KV}_1$, and $\mathrm{Exec}(\mathcal{C}, I_0)$ produces the crops (\cref{fig:supp_exec_code})}}\\[2pt]
\textit{Input:} \;\; \pillS{\texttt{[s\_1][s\_2][s\_3][s\_4]}} $\;\oplus\;$ \pillUser{\texttt{<image>} $\mathbf{V}_0$ \texttt{</image> What is written on the tag of the red backpack?}}\\[2.5pt]
\textit{Output:}\\
\begin{tcolorbox}[colback=pillObg, colframe=pillObd, boxrule=0.45pt, arc=1.5pt, left=4pt, right=4pt, top=2pt, bottom=2pt, boxsep=0pt, before skip=2pt, after skip=3pt]
\scriptsize
\texttt{Rationale: Two red bags sit near the bench (lower left and lower right), and their tags are too small to read in the global view. In Turn 2, inspect both zoomed crops to identify the backpack and transcribe its tag. In Turn 3, report the text from the confirmed backpack.}\\[1pt]
\texttt{<code>crops = [crop(I\_0, box=[184,512,276,635], z=4.0), crop(I\_0, box=[640,488,735,610], z=4.0)]</code><eos>}
\end{tcolorbox}
\hrule\vspace{3.5pt}
\textbf{2. Turn 2 (\emph{Inspect}, $P=2$ parallel batch branches):} \quad {\scriptsize\textcolor{slateheader}{Each branch reuses $\text{KV}_1$ and caches its appended crop + finding as $\text{KV}_2^{(p)}$}}\\[2pt]
\textit{Branch $p{=}1$:} \; \pillKV{$\text{KV}_1$} $\oplus$ \pillRun{\texttt{Crop 1: <image>} $\mathbf{V}^{(1)}$ \texttt{</image>}} $\longrightarrow$ \pillOut{\texttt{Finding 1: Red tote bag with no luggage tag.<eos>}}\\[2pt]
\textit{Branch $p{=}2$:} \; \pillKV{$\text{KV}_1$} $\oplus$ \pillRun{\texttt{Crop 2: <image>} $\mathbf{V}^{(2)}$ \texttt{</image>}} $\longrightarrow$ \pillOut{\texttt{Finding 2: Red backpack; white tag reads "ALPINE".<eos>}}
\vspace{3pt}\hrule\vspace{3.5pt}
\textbf{3. Turn 3 (\emph{Aggregate}):} \quad {\scriptsize\textcolor{slateheader}{Concatenate $\text{KV}_1$ with $[\text{KV}_2^{(1)}, \text{KV}_2^{(2)}]$ and append the aggregation cue}}\\[2pt]
\textit{Input:} \;\; \pillKV{$\text{KV}_1$} $\;\oplus\;$ \pillKV{$\text{KV}_2^{(1)}$} $\;\oplus\;$ \pillKV{$\text{KV}_2^{(2)}$} $\;\oplus\;$ \pillRun{\texttt{Final answer:}}\\[2pt]
\textit{Output:} \; \pillOut{\texttt{Crop 1 has no tag; the backpack in crop 2 reads "ALPINE". <answer>ALPINE</answer><eos>}}
\end{tcolorbox}
\vspace{-1mm}
\caption{\textbf{Example of Three-Turn Inference Token Sequences.} We show the input tokens, reused KV-caches ($\text{KV}_1, \text{KV}_2^{(p)}$), and VLM outputs across Turns~1--3 on an example with $P=2$ candidate crops ($\oplus$ denotes sequence concatenation; helper functions inside \texttt{<code>...</code>} are omitted for brevity, see \cref{fig:supp_exec_code}). For two-image tasks (\eg, VDiff-Bench-Gen), Turn~1 encodes both global images and each Turn~2 branch receives an aligned crop pair.}
\label{fig:supp_prompts}
\vspace{-1ex}
\end{figure*}

Pure RL (\eg, DeepEyes~\cite{zheng2025deepeyes}) avoids trajectory annotation by training only on final-answer rewards, yet struggles to cold-start complex tool or code use beyond basic cropping. Recent methods such as Pixel-Reasoner~\cite{su2025pixelreasoner}, DeepEyesV2~\cite{hong2025deepeyesv2}, and CodeV~\cite{hou2026codev} therefore adopt a cold-start SFT stage over 8.4K--253K synthesized trajectories prior to RL. Existing SFT+RL search pipelines rely solely on LLMs to synthesize training trajectories, as manually annotating variable-turn sequential search is impractical due to search path ambiguity and human bias (\eg, multiple visiting orders or zoom-in paths can reach the same target, causing annotations to vary across people). Without human demonstrations, the quality of LLM-synthesized trajectories is difficult to guarantee. In contrast, our deterministic three-turn trajectory proposes all candidate regions upfront in Turn~1 with no specific order required, making human annotation of region proposals and local findings straightforward and low-bias. Using these human-annotated seeds (140 across five tasks) as in-context demonstrations guides the LLM to produce consistent, high-quality trajectories for SFT.

\section{Turn-Level Token Structure and Inference Workflow}
\label{supp:sec_prompts}

\cref{fig:supp_prompts} provides a concrete example of the inference token sequences at each turn. The user only provides the input image $I_0$ and text query $\mathbf{q}$ in Turn~1, prepended with the $K=4$ learned prefix tokens $\mathbf{S}$ and no system prompt. From this input, the VLM generates the rationale $r$ and the Python cropping script $\mathcal{C}$ (helper functions omitted for brevity; see \cref{fig:supp_exec_code}). The Turn~1 input and output are cached together as $\text{KV}_1$, and executing $\mathrm{Exec}(\mathcal{C}, I_0)$ produces the candidate crops $\{I^{(p)}\}_{p=1}^P$. In Turn~2, each parallel branch $p \in \{1, \dots, P\}$ reuses $\text{KV}_1$, appends \texttt{Crop $p$:} with the encoded patch $\mathbf{V}^{(p)}$ (indexing $p$ so each branch can follow region-specific instructions in $r$), and generates its regional finding $\mathbf{a}^{(p)}$, caching the appended crop and finding as $\text{KV}_2^{(p)}$. In Turn~3, $\text{KV}_1$ is concatenated with $[\text{KV}_2^{(1)}, \dots, \text{KV}_2^{(P)}]$ followed by \texttt{Final answer:}, and the VLM aggregates the regional findings to output the final answer $\mathbf{y}^*$.

\cref{fig:supp_baseline_prompt} shows the system prompt used for the text-prompting baseline in \cref{subsec:ablation_studies}. In this ablation, we remove the learned prefix tokens $\mathbf{S}$ in Turn~1 and prepend explicit text instructions describing the three-turn workflow, while keeping the rest of the inference pipeline unchanged.

\section{Visualizations of Three-Turn Search Trajectories}
\label{supp:sec_visualizations}
\begin{figure*}[t]
\centering
\begin{tcolorbox}[colback=white, colframe=promptframe, boxrule=0.7pt, arc=2pt, left=5pt, right=5pt, top=5pt, bottom=5pt, title={\textbf{\footnotesize Scenario 1 (Cluttered Camouflaged Search, WaldoBench)}}]
\footnotesize
\textbf{User Query $\mathbf{q}$:} \texttt{Where is Waldo? Return his location as a point.}\\[3pt]
\includegraphics[width=\linewidth]{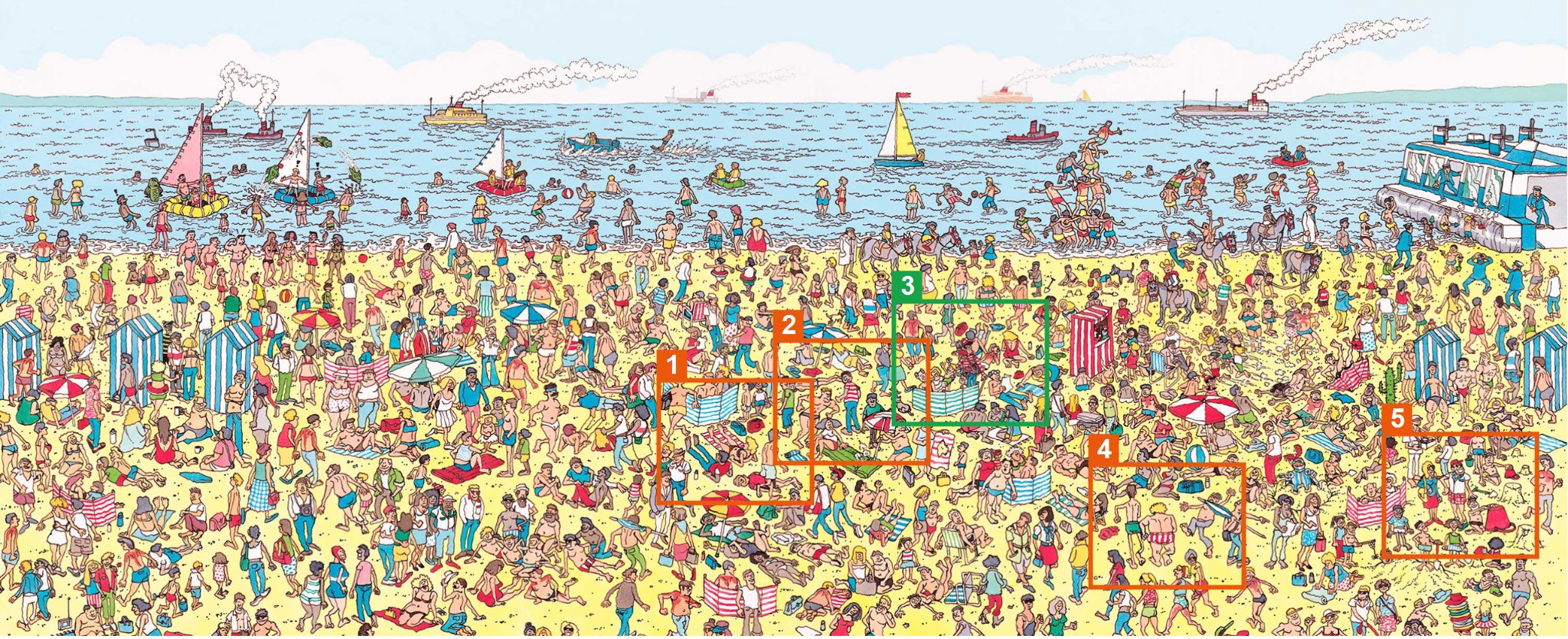}\\[3pt]
\textbf{Turn 1 (\emph{Propose}) Output $(r, \mathcal{C})$:}\\
\texttt{Rationale:} The image is a crowded beach with hundreds of small figures. Waldo wears a red-and-white striped shirt, a bobble hat, and round glasses, and at this scale he covers only a few dozen pixels. Red-and-white patterns that could be him appear in five places on the sand, so I zoom into all of them. In Turn~2, check each crop for the full outfit (stripes, hat, and glasses) and reject striped towels, windbreaks, swimwear, and umbrellas. In Turn~3, keep the crop that matches and map its location back to the full image.\\[2pt]
\scriptsize
\texttt{Code: crops = [crop(I\_0, box=[420, 595, 518, 787], z=3.0), crop(I\_0, box=[494, 532, 592, 724], z=3.0),}\\
\texttt{\phantom{Code: crops = [}crop(I\_0, box=[570, 471, 667, 663], z=3.0), crop(I\_0, box=[695, 727, 793, 918], z=3.0),}\\
\texttt{\phantom{Code: crops = [}crop(I\_0, box=[882, 679, 980, 871], z=3.0)]}\\[4pt]
\footnotesize
\hrule\vspace{4pt}
\textbf{Turn 2 (\emph{Inspect}): parallel zoomed crops and findings $\{\mathbf{a}^{(p)}\}_{p=1}^{5}$}\\[3pt]
\begin{minipage}[t]{0.19\linewidth}\centering
\includegraphics[width=\linewidth]{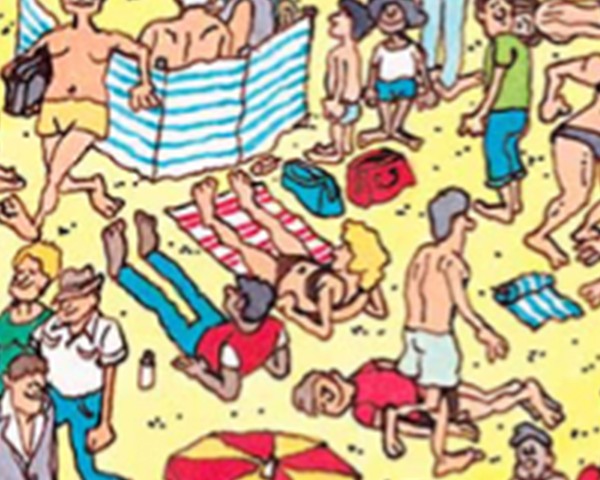}\\[1pt]
\scriptsize\raggedright \textbf{Finding 1:} Not found. The red-and-white stripes are a beach towel under a sunbather.
\end{minipage}\hfill
\begin{minipage}[t]{0.19\linewidth}\centering
\includegraphics[width=\linewidth]{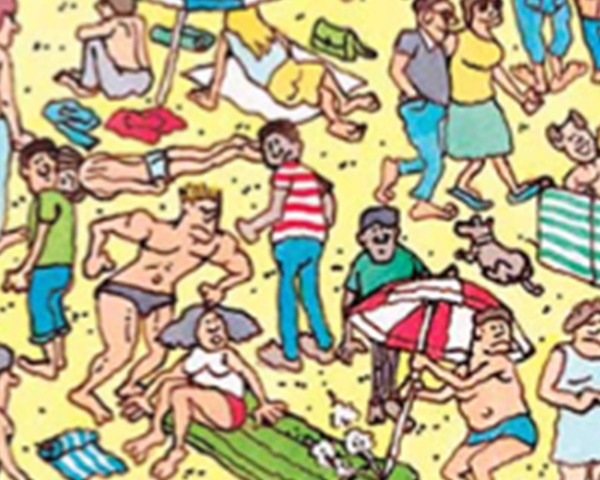}\\[1pt]
\scriptsize\raggedright \textbf{Finding 2:} Not found. A man wears a striped shirt, but he has no hat or glasses.
\end{minipage}\hfill
\begin{minipage}[t]{0.19\linewidth}\centering
\includegraphics[width=\linewidth]{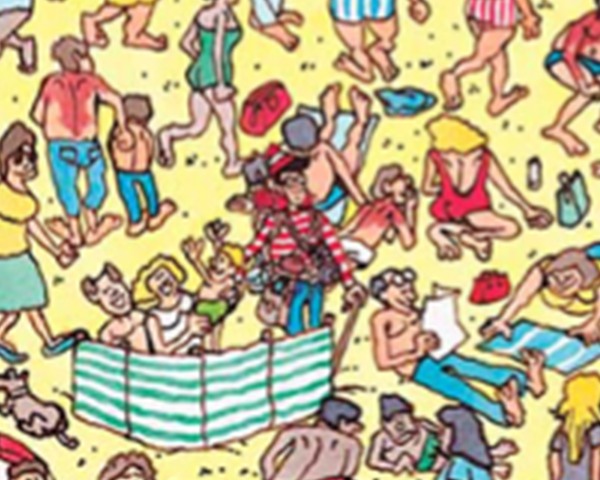}\\[1pt]
\scriptsize\raggedright \textbf{Finding 3:} \textbf{Found.} A figure behind the green windbreak has the striped shirt, bobble hat, and glasses, at about [0.50, 0.50] in this crop.
\end{minipage}\hfill
\begin{minipage}[t]{0.19\linewidth}\centering
\includegraphics[width=\linewidth]{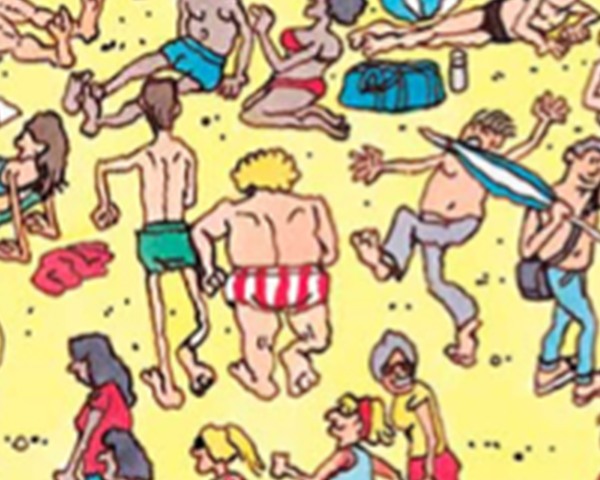}\\[1pt]
\scriptsize\raggedright \textbf{Finding 4:} Not found. The stripes are the swim shorts of a man seen from behind.
\end{minipage}\hfill
\begin{minipage}[t]{0.19\linewidth}\centering
\includegraphics[width=\linewidth]{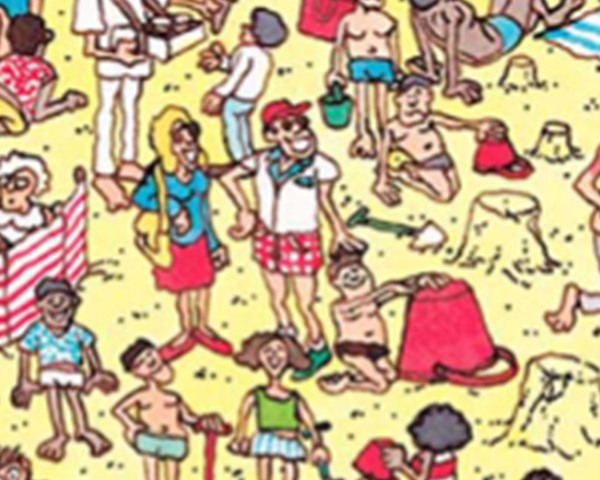}\\[1pt]
\scriptsize\raggedright \textbf{Finding 5:} Not found. A man has a red cap and checkered shorts, but a plain white shirt.
\end{minipage}\\[4pt]
\hrule\vspace{4pt}
\textbf{Turn 3 (\emph{Aggregate}) Final Answer $\mathbf{y}^*$:} Only crop 3 contains a figure with the full outfit; crops 1, 2, 4, and 5 are look-alike distractors. Mapping the crop location back to the full image: \texttt{Waldo is at [619, 567].}
\end{tcolorbox}
\caption{\textbf{Scenario 1: Camouflaged Target Search in Extreme Clutter (WaldoBench~\cite{waldobench2026}).} Turn~1 proposes all plausible red-and-white striped candidate regions at once (boxes 1--5 on the global view; the target is shown in green and distractors in orange). Turn~2 inspects the zoomed crops in parallel and rules out look-alike distractors, and Turn~3 aggregates the findings into the final location in normalized $[0, 1000]$ coordinates.}
\label{fig:supp_traj_waldo}
\end{figure*}

\begin{figure*}[t]
\centering
\begin{tcolorbox}[colback=white, colframe=promptframe, boxrule=0.7pt, arc=2pt, left=5pt, right=5pt, top=5pt, bottom=5pt, title={\textbf{\footnotesize Scenario 2 (Dense Object Counting, UNICBench \textit{Hard})}}]
\footnotesize
\begin{minipage}[c]{0.30\linewidth}
\includegraphics[width=\linewidth]{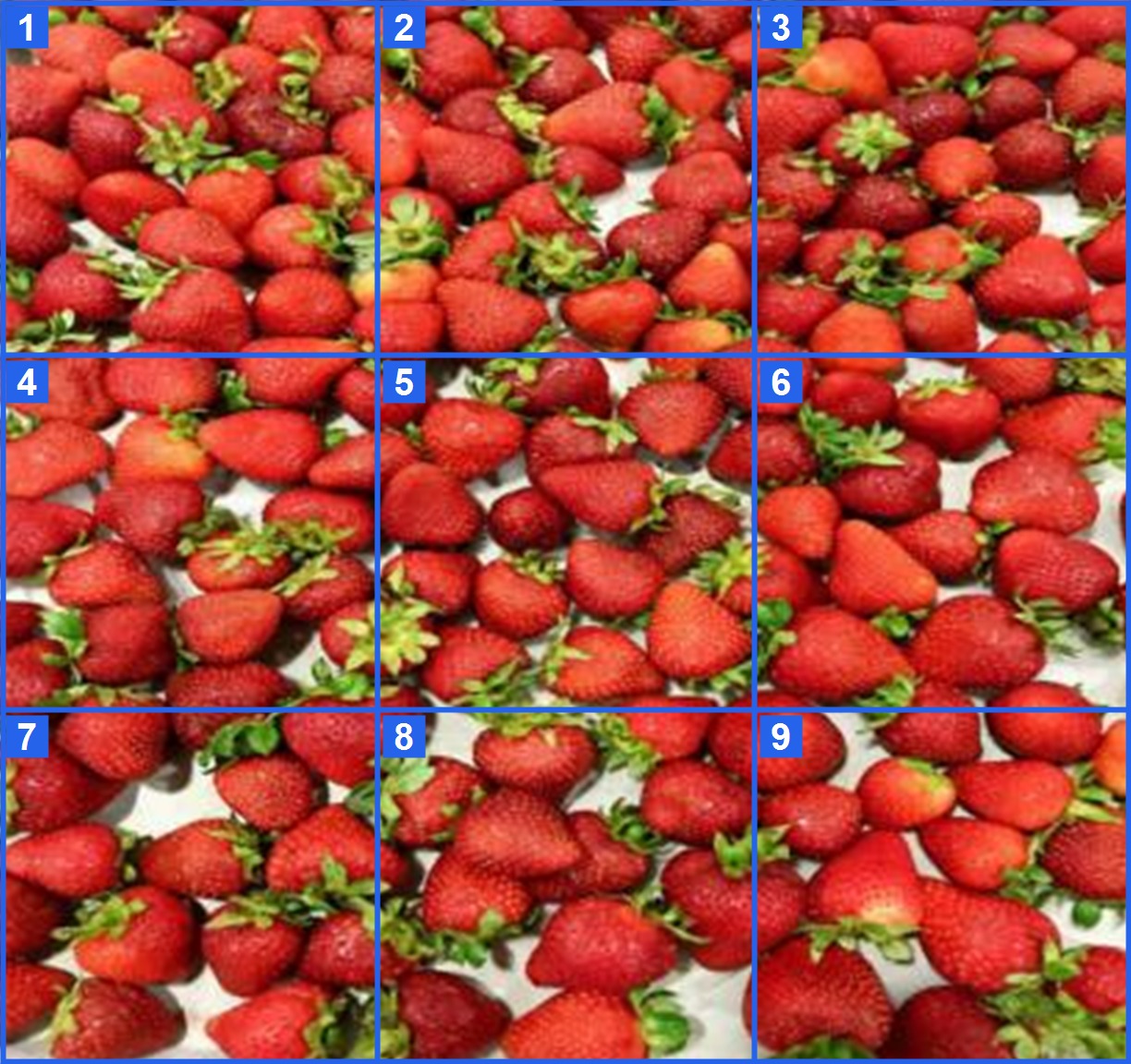}
\end{minipage}\hfill
\begin{minipage}[c]{0.67\linewidth}
\textbf{User Query $\mathbf{q}$:} \texttt{How many strawberries are there in the image?}\\[4pt]
\textbf{Turn 1 (\emph{Propose}) Output $(r, \mathcal{C})$:}\\
\texttt{Rationale:} The image is densely and almost uniformly covered with strawberries, far too many to count reliably in a single pass, and there is no sparse region that can be skipped. I therefore split the image into a regular $3 \times 3$ grid and zoom into every tile. In Turn~2, count the strawberries in each tile; a strawberry cut by a tile border is counted only in the tile that contains its center, so that no fruit is counted twice. In Turn~3, sum the nine sub-counts to obtain the total.\\[2pt]
\texttt{Code: crops = grid(I\_0, rows=3, cols=3, z=3.0)}
\end{minipage}\\[4pt]
\hrule\vspace{4pt}
\textbf{Turn 2 (\emph{Inspect}): parallel sub-counts $\{\mathbf{a}^{(p)}\}_{p=1}^{9}$}\\[3pt]
\begin{minipage}[t]{0.102\linewidth}\centering \includegraphics[width=\linewidth]{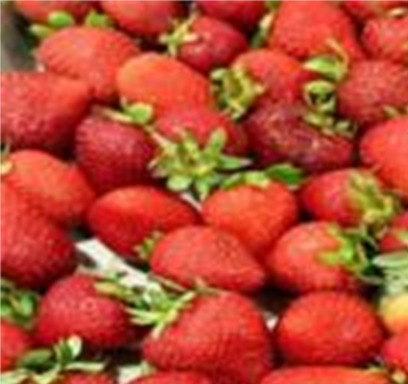}\\[1pt] \scriptsize \textbf{Tile 1:} 24 \end{minipage}\hfill
\begin{minipage}[t]{0.102\linewidth}\centering \includegraphics[width=\linewidth]{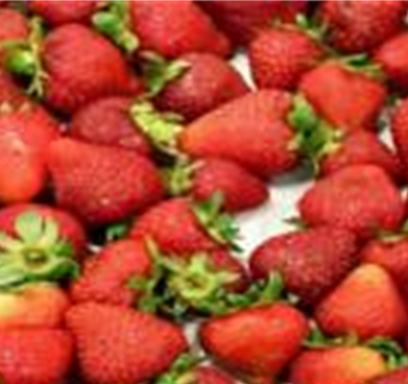}\\[1pt] \scriptsize \textbf{Tile 2:} 27 \end{minipage}\hfill
\begin{minipage}[t]{0.102\linewidth}\centering \includegraphics[width=\linewidth]{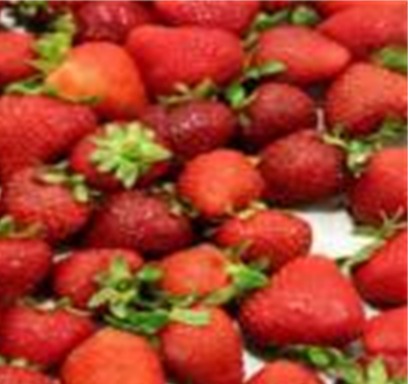}\\[1pt] \scriptsize \textbf{Tile 3:} 26 \end{minipage}\hfill
\begin{minipage}[t]{0.102\linewidth}\centering \includegraphics[width=\linewidth]{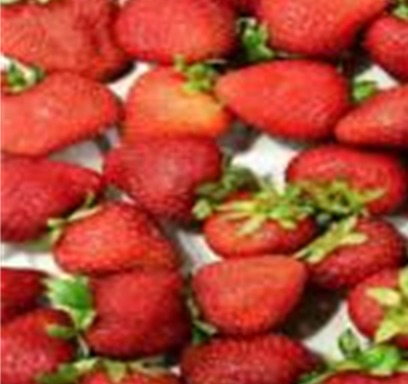}\\[1pt] \scriptsize \textbf{Tile 4:} 18 \end{minipage}\hfill
\begin{minipage}[t]{0.102\linewidth}\centering \includegraphics[width=\linewidth]{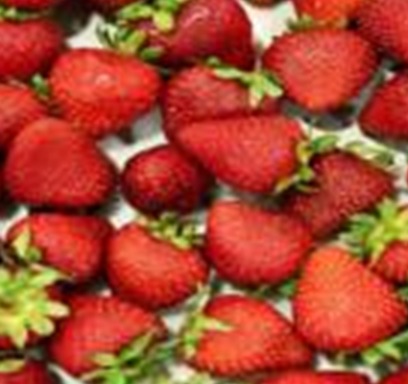}\\[1pt] \scriptsize \textbf{Tile 5:} 16 \end{minipage}\hfill
\begin{minipage}[t]{0.102\linewidth}\centering \includegraphics[width=\linewidth]{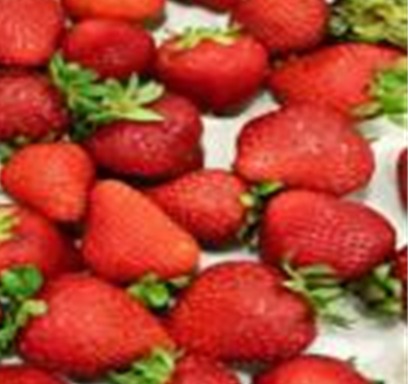}\\[1pt] \scriptsize \textbf{Tile 6:} 15 \end{minipage}\hfill
\begin{minipage}[t]{0.102\linewidth}\centering \includegraphics[width=\linewidth]{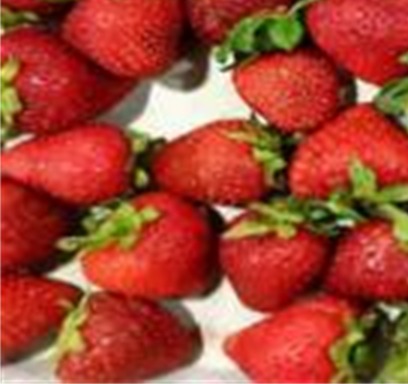}\\[1pt] \scriptsize \textbf{Tile 7:} 14 \end{minipage}\hfill
\begin{minipage}[t]{0.102\linewidth}\centering \includegraphics[width=\linewidth]{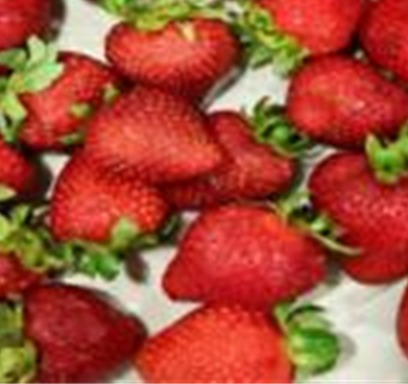}\\[1pt] \scriptsize \textbf{Tile 8:} 11 \end{minipage}\hfill
\begin{minipage}[t]{0.102\linewidth}\centering \includegraphics[width=\linewidth]{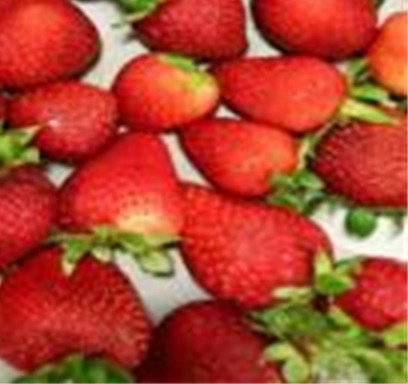}\\[1pt] \scriptsize \textbf{Tile 9:} 14 \end{minipage}\\[4pt]
\hrule\vspace{4pt}
\textbf{Turn 3 (\emph{Aggregate}) Final Answer $\mathbf{y}^*$:} $24 + 27 + 26 + 18 + 16 + 15 + 14 + 11 + 14 = 165$. Border strawberries were assigned to a single tile, so no correction is needed. \texttt{There are 165 strawberries.} (Ground truth: 165.)
\end{tcolorbox}
\caption{\textbf{Scenario 2: Dense Object Counting (UNICBench \textit{Hard}~\cite{unicbench2026}).} Single-pass VLMs under-count such scenes since individual instances merge at the global resolution. Turn~1 partitions the image into a regular grid with one line of code, Turn~2 counts each zoomed tile in parallel under a shared rule for border instances, and Turn~3 sums the sub-counts into the final answer.}
\label{fig:supp_traj_counting}
\end{figure*}

\begin{figure*}[t]
\centering
\begin{tcolorbox}[colback=white, colframe=promptframe, boxrule=0.7pt, arc=2pt, left=5pt, right=5pt, top=5pt, bottom=5pt, title={\textbf{\footnotesize Scenario 3 (Local-to-Local Relational Search: A Clue in One Region Determines the Target in Another)}}]
\footnotesize
\begin{minipage}[c]{0.50\linewidth}
\includegraphics[width=\linewidth]{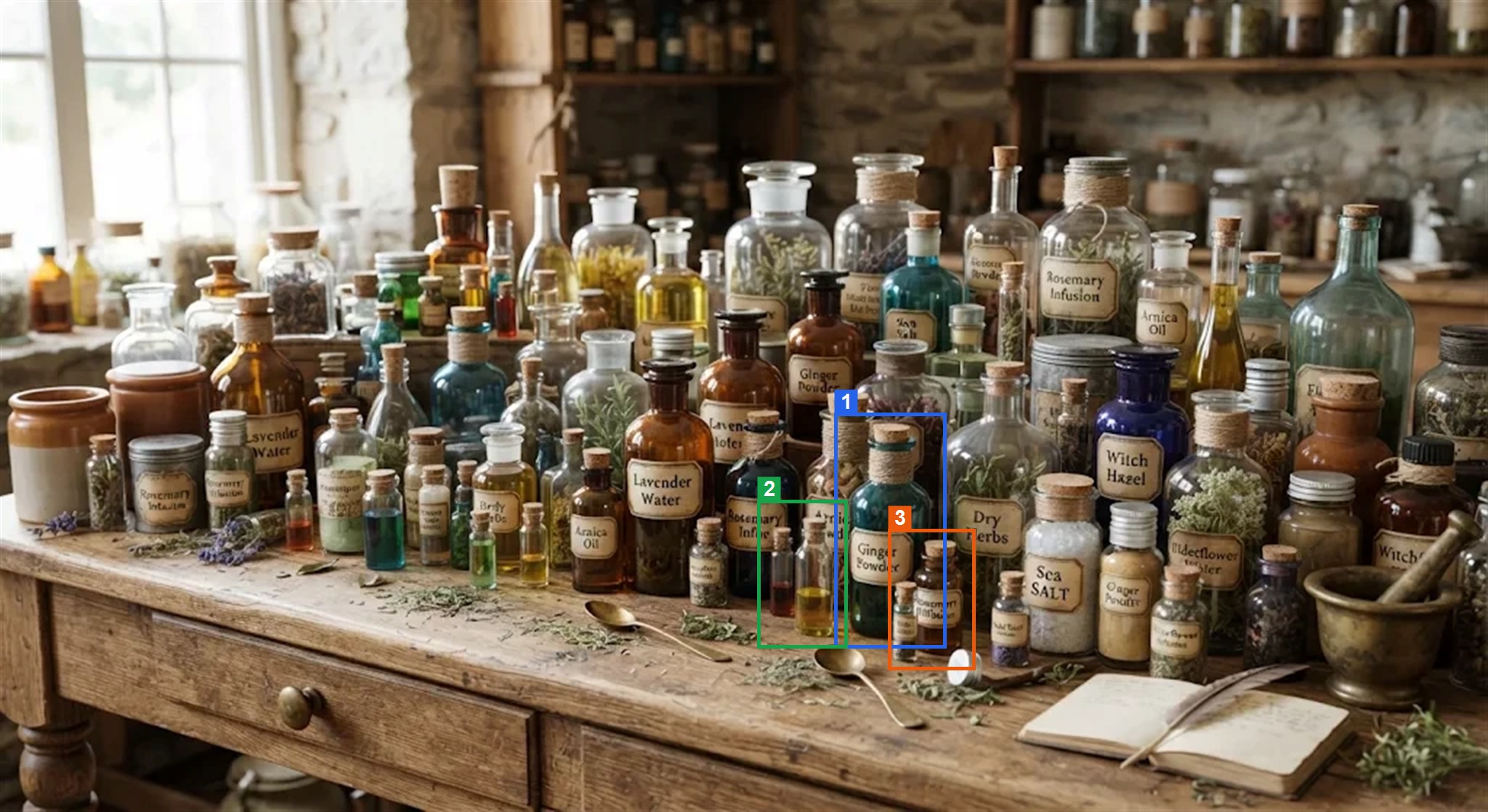}
\end{minipage}\hfill
\begin{minipage}[c]{0.47\linewidth}
\textbf{User Query $\mathbf{q}$:} \texttt{What color is the liquid in the small transparent bottle next to the Ginger Powder bottle?}\\[4pt]
\textbf{Turn 1 (\emph{Propose}) Output $(r, \mathcal{C})$:}\\
\texttt{Rationale:} The table holds dozens of labeled bottles, and the labels are too small to read reliably in the global view. A teal bottle in the front row appears to carry the ``Ginger Powder'' label, with small bottles on both of its sides. I crop the labeled bottle to confirm the clue (crop~1) and the small bottles to its left (crop~2) and right (crop~3) as candidates. In Turn~2, read the label in crop~1, and for crops~2 and~3 report whether each small bottle is transparent and what it contains. In Turn~3, pick the transparent bottle adjacent to the confirmed bottle and report its liquid color.\\[2pt]
{\scriptsize
\texttt{Code: crops = [}\\
\texttt{\phantom{Co}crop(I\_0, box=[561, 510, 635, 796], z=5.0),}\\
\texttt{\phantom{Co}crop(I\_0, box=[510, 617, 568, 796], z=6.0),}\\
\texttt{\phantom{Co}crop(I\_0, box=[598, 653, 654, 823], z=6.0)]}}
\end{minipage}\\[4pt]
\hrule\vspace{4pt}
\textbf{Turn 2 (\emph{Inspect}): parallel zoomed crops and findings $\{\mathbf{a}^{(p)}\}_{p=1}^{3}$}\\[3pt]
\begin{minipage}[c]{0.32\linewidth}
\begin{minipage}[c]{0.27\linewidth}\includegraphics[width=\linewidth]{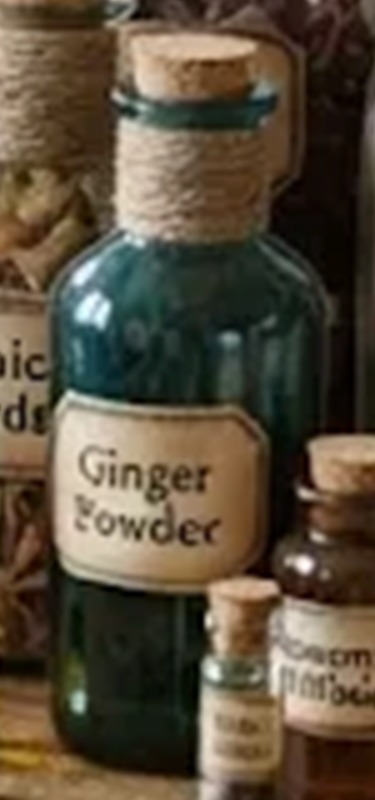}\end{minipage}\hfill
\begin{minipage}[c]{0.69\linewidth}\scriptsize\raggedright \textbf{Finding 1 (clue):} A teal glass bottle with a twine-wrapped neck. Its label reads ``Ginger Powder''. This is the reference bottle.\end{minipage}
\end{minipage}\hfill
\begin{minipage}[c]{0.32\linewidth}
\begin{minipage}[c]{0.34\linewidth}\includegraphics[width=\linewidth]{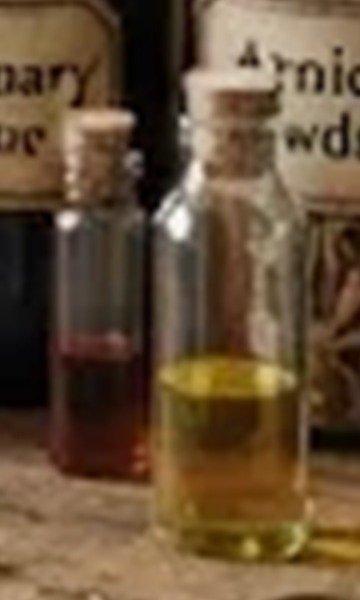}\end{minipage}\hfill
\begin{minipage}[c]{0.62\linewidth}\scriptsize\raggedright \textbf{Finding 2:} Two small clear glass bottles with cork stoppers. The larger one, directly beside the teal bottle, holds a \textbf{yellow} liquid; the thinner vial further left holds a red liquid.\end{minipage}
\end{minipage}\hfill
\begin{minipage}[c]{0.32\linewidth}
\begin{minipage}[c]{0.34\linewidth}\includegraphics[width=\linewidth]{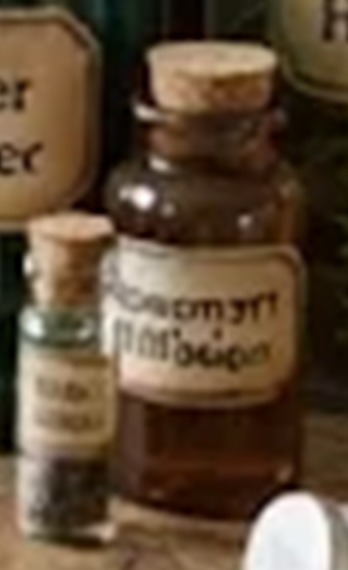}\end{minipage}\hfill
\begin{minipage}[c]{0.62\linewidth}\scriptsize\raggedright \textbf{Finding 3:} Two small labeled bottles. The larger one is brown glass, not transparent; the smaller one holds dried herbs, not a liquid.\end{minipage}
\end{minipage}\\[4pt]
\hrule\vspace{4pt}
\textbf{Turn 3 (\emph{Aggregate}) Final Answer $\mathbf{y}^*$:} Crop~1 confirms the Ginger Powder bottle. The bottles on its right (crop~3) are not transparent bottles of liquid, while the clear bottle directly on its left (crop~2) holds a yellow liquid. \texttt{The liquid is yellow.}
\end{tcolorbox}
\caption{\textbf{Scenario 3: Local-to-Local Relational Search.} Answering the query requires reading a fine-grained clue in one region (the label in crop~1) to decide which neighboring region contains the target (crops~2--3). Although the label is illegible in the downsampled global view, the \emph{locations} of both the clue-bearing bottle and its neighbors are visible in the global layout. Turn~1 therefore proposes crops for the clue (blue) and all candidates (green: target, orange: distractor) at once, Turn~2 inspects them in parallel, and Turn~3 resolves the relation without an additional search step.}
\label{fig:supp_traj_multiclue}
\end{figure*}

\begin{figure*}[t]
\centering
\begin{tcolorbox}[colback=white, colframe=promptframe, boxrule=0.7pt, arc=2pt, left=5pt, right=5pt, top=5pt, bottom=5pt, title={\textbf{\footnotesize Scenario 4 (Ultra-High-Resolution Perception, HR-Bench 8K)}}]
\footnotesize
\begin{minipage}[c]{0.44\linewidth}
\includegraphics[width=\linewidth]{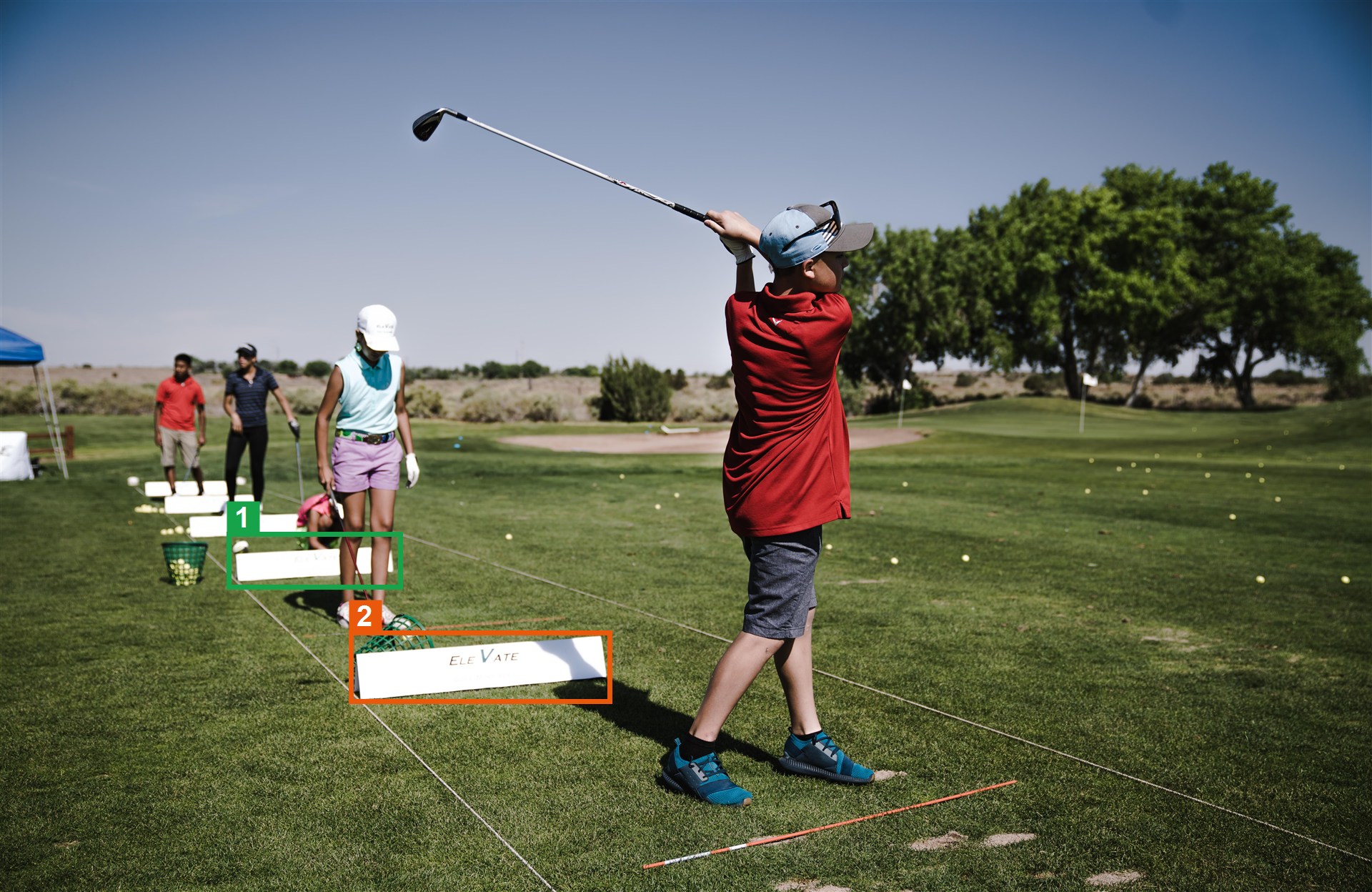}
\end{minipage}\hfill
\begin{minipage}[c]{0.53\linewidth}
\textbf{User Query $\mathbf{q}$:} \texttt{What is written on the white board in front of the girl in purple shorts? (A) ELEVATE (B) ELEVATE. (C) ELVTE (D) CeVaY}\\[4pt]
\textbf{Turn 1 (\emph{Propose}) Output $(r, \mathcal{C})$:}\\
\texttt{Rationale:} The image is $7680 \times 4992$, and the lettering on the boards is far too small to read in the global view. The girl in purple shorts stands on the left, with a white board at her feet; a similar board lies in the foreground. I crop the board in front of the girl at native resolution (crop~1) and the foreground board as a reference for the same sign (crop~2). In Turn~2, transcribe the text on each board. In Turn~3, answer with the text on the girl's board, using the reference board only to resolve unclear letters.\\[2pt]
{\scriptsize
\texttt{Code: crops = [}\\
\texttt{\phantom{Co}crop(I\_0, box=[167, 598, 292, 658], z=1.0),}\\
\texttt{\phantom{Co}crop(I\_0, box=[256, 709, 444, 786], z=1.0)]}}
\end{minipage}\\[4pt]
\hrule\vspace{4pt}
\textbf{Turn 2 (\emph{Inspect}): native-resolution crops and findings}\\[3pt]
\begin{minipage}[t]{0.32\linewidth}\centering
\includegraphics[width=\linewidth]{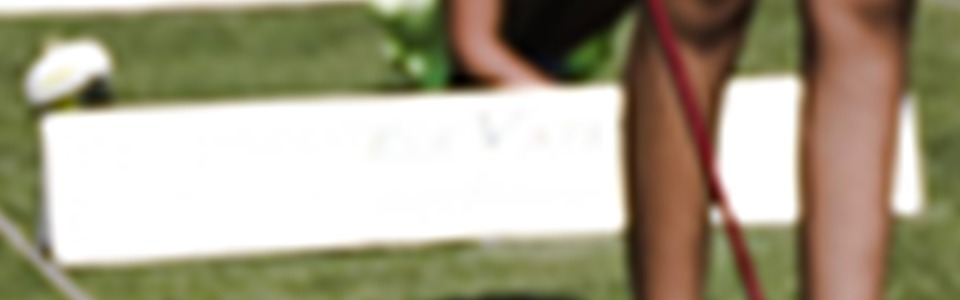}\\[1pt]
\scriptsize\raggedright \textbf{Same region in the global view $\mathbf{V}_0$} (for reference): the lettering is not resolved.
\end{minipage}\hfill
\begin{minipage}[t]{0.32\linewidth}\centering
\includegraphics[width=\linewidth]{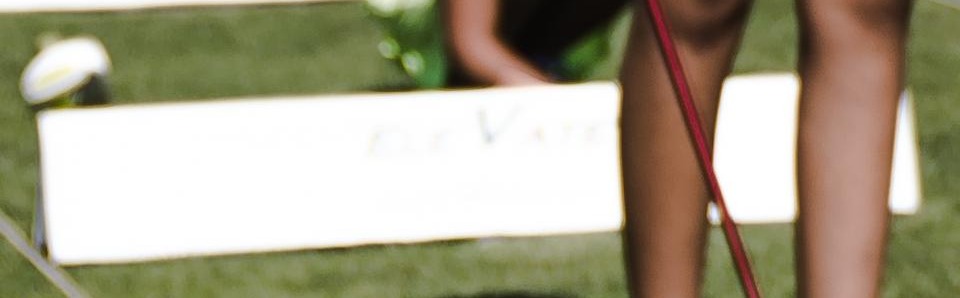}\\[1pt]
\scriptsize\raggedright \textbf{Finding 1:} The board at the girl's feet carries faint, out-of-focus lettering with a large ``V'' in the middle, consistent with ``ELEVATE''; it is partly hidden by her legs.
\end{minipage}\hfill
\begin{minipage}[t]{0.32\linewidth}\centering
\includegraphics[width=\linewidth]{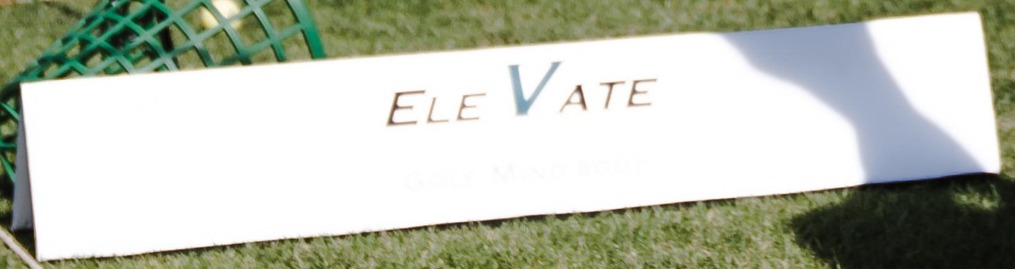}\\[1pt]
\scriptsize\raggedright \textbf{Finding 2:} The foreground board is the same sign and is in focus. It reads ``ELEVATE'', with an enlarged ``V'' and no trailing period.
\end{minipage}\\[4pt]
\hrule\vspace{4pt}
\textbf{Turn 3 (\emph{Aggregate}) Final Answer $\mathbf{y}^*$:} The girl's board shows the same lettering as the in-focus reference board, which reads ``ELEVATE'' without a period: \texttt{(A) ELEVATE}.
\end{tcolorbox}
\caption{\textbf{Scenario 4: Ultra-High-Resolution Perception at 8K Scale (HR-Bench 8K~\cite{wang2024hrbench}).} More than 38M pixels are compressed into the global view $\mathbf{V}_0$, where the queried text is not resolved (bottom left). Turn~1 uses the coarse scene layout to propose crops at native resolution, Turn~2 reads them in parallel, and Turn~3 combines a partially legible target (green) with an in-focus reference (orange) to select the answer.}
\label{fig:supp_traj_hrbench}
\end{figure*}

\begin{figure*}[t]
\centering
\begin{tcolorbox}[colback=white, colframe=promptframe, boxrule=0.7pt, arc=2pt, left=5pt, right=5pt, top=5pt, bottom=5pt, title={\textbf{\footnotesize Scenario 5 (Multi-Image Difference Detection, VDiff-Bench-Gen)}}]
\footnotesize
\textbf{User Query $\mathbf{q}$:} \texttt{List all differences between the two images.}\\[3pt]
\begin{minipage}[c]{0.495\linewidth}\centering
\includegraphics[width=\linewidth]{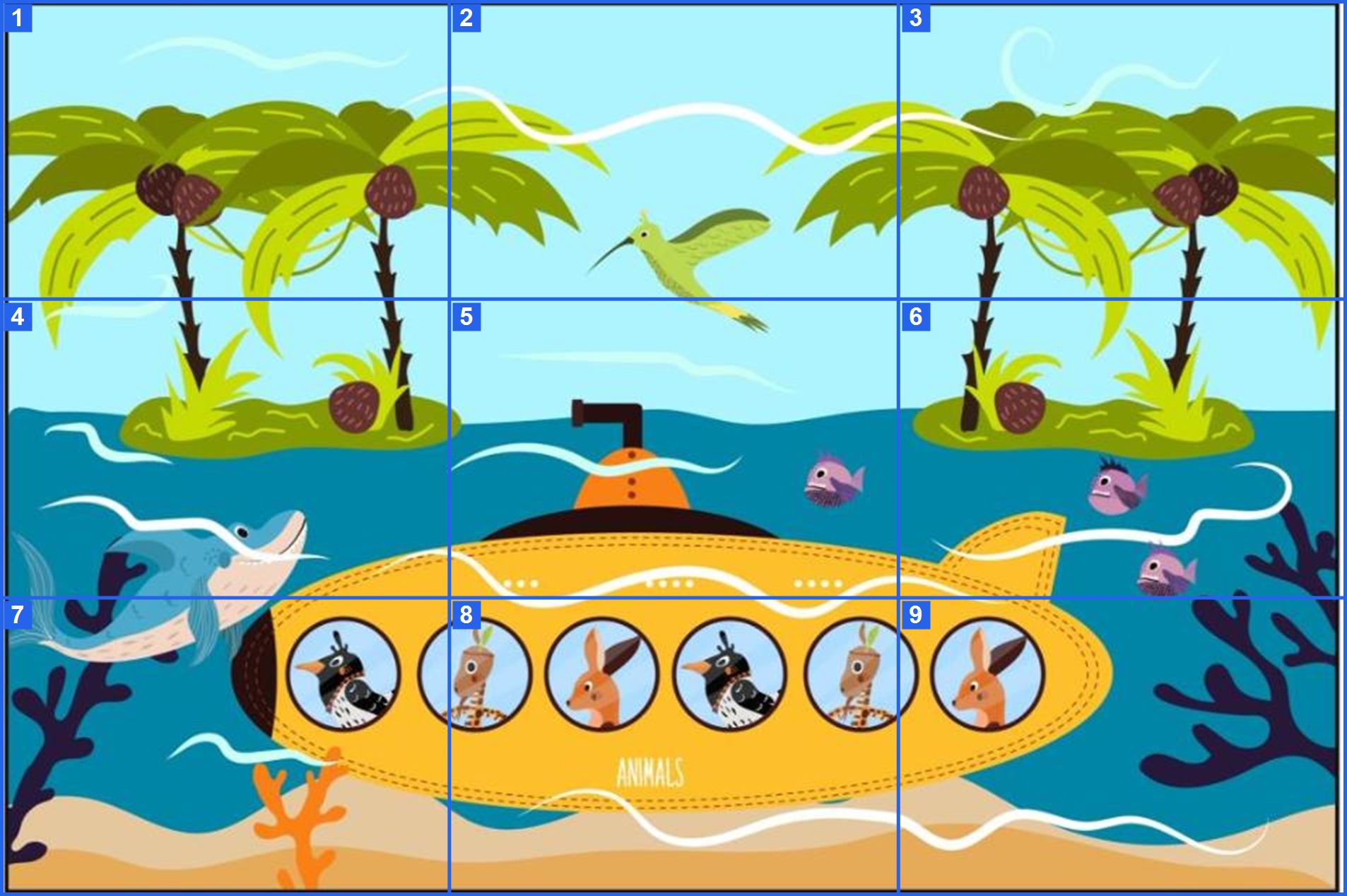}\\[1pt] \scriptsize Image 1
\end{minipage}\hfill
\begin{minipage}[c]{0.495\linewidth}\centering
\includegraphics[width=\linewidth]{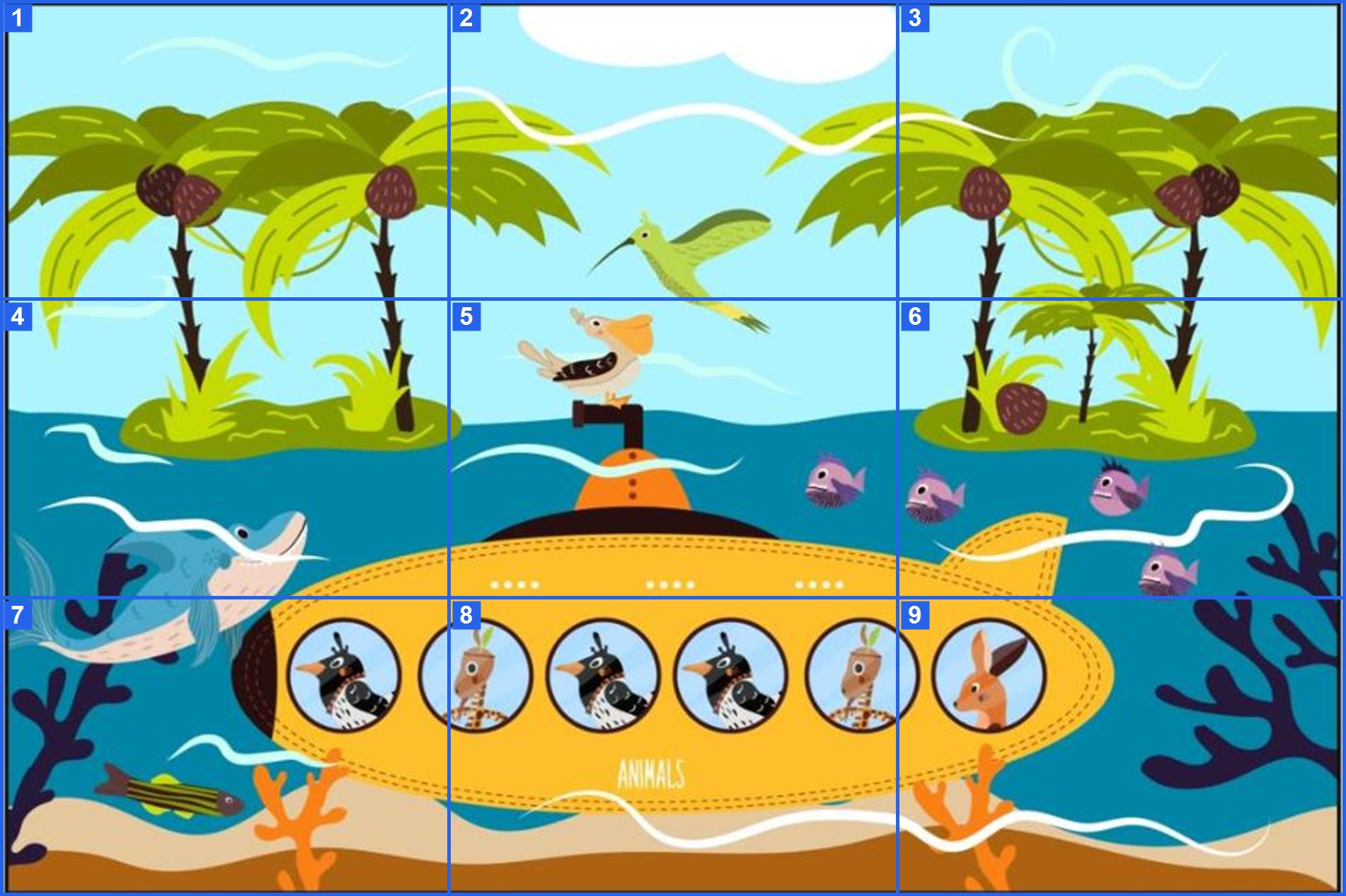}\\[1pt] \scriptsize Image 2
\end{minipage}\\[4pt]
\textbf{Turn 1 (\emph{Propose}) Output $(r, \mathcal{C})$:}\\
\texttt{Rationale:} The two images show the same cartoon scene, and the differences may be anywhere, so no region can be skipped. I split both images with the same $3 \times 3$ grid, so that tile $k$ of image~1 and tile $k$ of image~2 cover the same area. In Turn~2, compare the two tiles of each pair and list every object that appears, disappears, or changes. In Turn~3, merge the per-tile lists and remove differences reported twice across neighboring tiles.\\[2pt]
{\scriptsize \texttt{Code: crops = list(zip(grid(I\_0, rows=3, cols=3, z=2.0), grid(I\_1, rows=3, cols=3, z=2.0)))}}\\[4pt]
\hrule\vspace{4pt}
\textbf{Turn 2 (\emph{Inspect}): aligned tile pairs and findings (3 of 9 pairs shown)}\\[3pt]
\begin{minipage}[t]{0.32\linewidth}\centering
\includegraphics[width=0.49\linewidth]{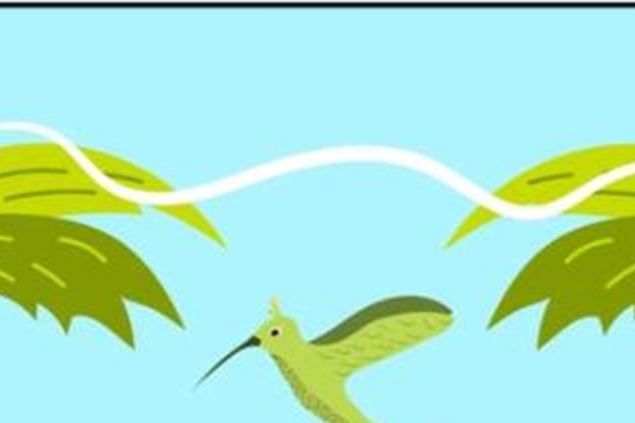}\hfill\includegraphics[width=0.49\linewidth]{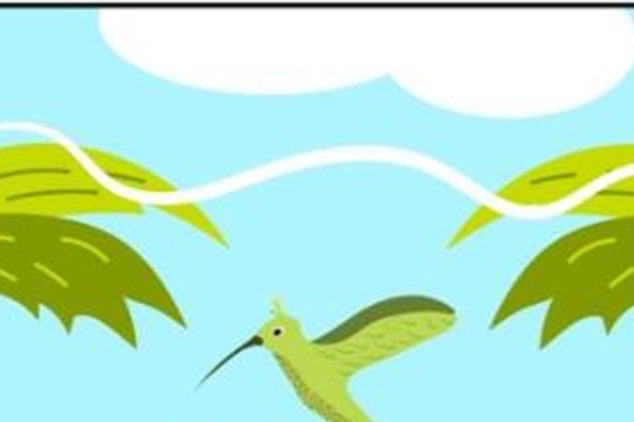}\\[1pt]
\scriptsize\raggedright \textbf{Pair 2:} A white cloud appears at the top of image~2.
\end{minipage}\hfill
\begin{minipage}[t]{0.32\linewidth}\centering
\includegraphics[width=0.49\linewidth]{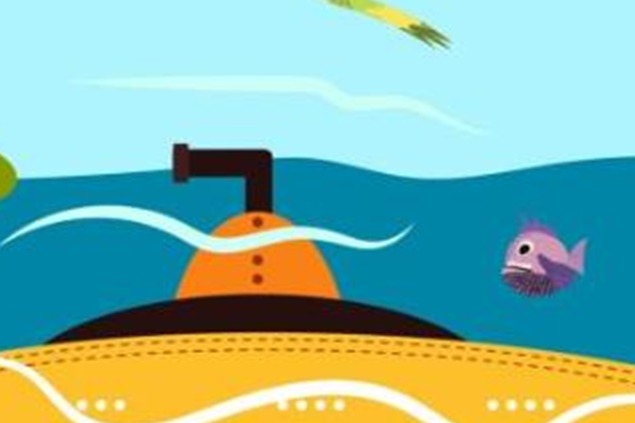}\hfill\includegraphics[width=0.49\linewidth]{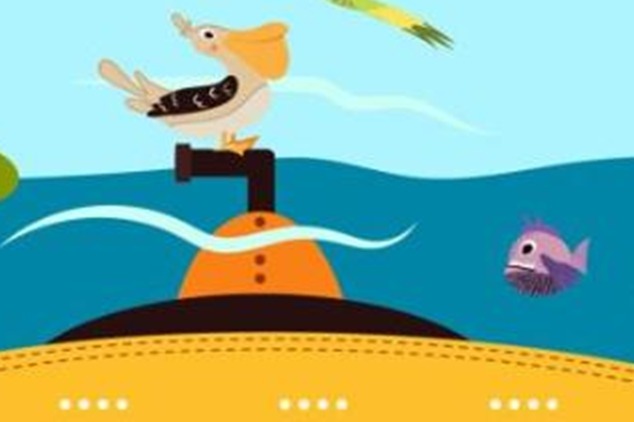}\\[1pt]
\scriptsize\raggedright \textbf{Pair 5:} A pelican appears on the periscope in image~2. The dot groups on the hull change from three to four dots. The white wave line across the top of the hull disappears.
\end{minipage}\hfill
\begin{minipage}[t]{0.32\linewidth}\centering
\includegraphics[width=0.49\linewidth]{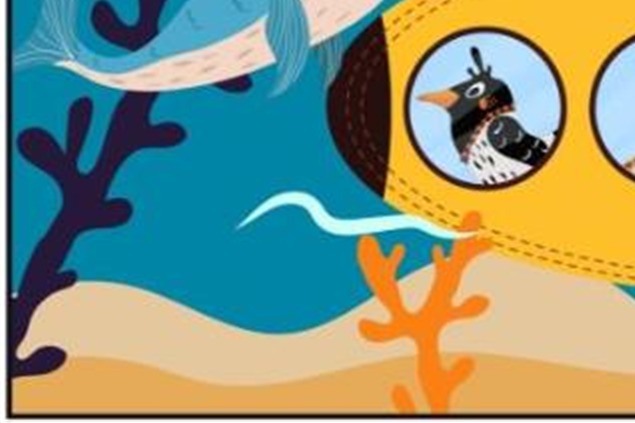}\hfill\includegraphics[width=0.49\linewidth]{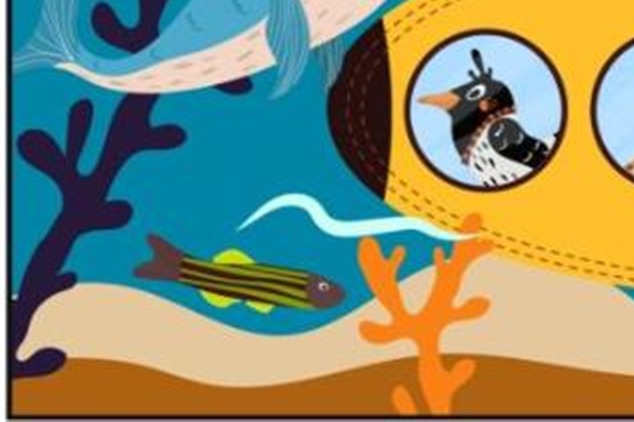}\\[1pt]
\scriptsize\raggedright \textbf{Pair 7:} A striped green fish appears in the lower left of image~2.
\end{minipage}\\[4pt]
\hrule\vspace{4pt}
\textbf{Turn 3 (\emph{Aggregate}) Final Answer $\mathbf{y}^*$:} \texttt{In image 2: (1) a cloud appears at the top; (2) a pelican appears on the periscope; (3) the hull dots change from three to four per group; (4) the white wave line across the hull disappears; (5) a striped green fish appears in the lower left; (6) a small palm tree appears on the right island; (7) the coconut at the foot of the left island disappears; (8) an extra purple fish appears; (9) the animal in the third window changes from a rabbit to a bird; (10) an orange coral appears in the lower right.}
\end{tcolorbox}
\caption{\textbf{Scenario 5: Fine-Grained Multi-Image Difference Detection (VDiff-Bench-Gen~\cite{vdiffbench2026}).} Turn~1 partitions both images with the same grid, so that corresponding tiles can be compared side by side. Turn~2 inspects each aligned tile pair in parallel and reports local differences, and Turn~3 merges them into a single list, removing duplicates reported by neighboring tiles.}
\label{fig:supp_traj_vdiff}
\end{figure*}

\cref{fig:supp_traj_waldo,fig:supp_traj_counting,fig:supp_traj_multiclue,fig:supp_traj_hrbench,fig:supp_traj_vdiff} show example three-turn trajectories across our five benchmarks (for readability, we display a subset of crops when $P$ is large, omit helper functions in $\mathcal{C}$, and shorten long outputs). In \cref{fig:supp_traj_waldo} (WaldoBench), the model zooms into five candidate regions to separate Waldo from look-alike distractors, while in \cref{fig:supp_traj_counting} (UNICBench) it splits a crowded scene into a $3{\times}3$ grid to count objects without double-counting on borders. In \cref{fig:supp_traj_multiclue} ($V^*$~Bench), reading a small label is needed to tell which neighboring bottle is the target; cropping the labeled bottle and its neighbors together in Turn~1 lets the model answer in Turn~3 without extra search turns. In \cref{fig:supp_traj_hrbench} (HR-Bench 8K), the model reads blurry text in an 8K photo by also cropping a clearer sign nearby, and in \cref{fig:supp_traj_vdiff} (VDiff-Bench-Gen) it compares matching tiles across two images to spot small differences. Together, these examples show that our proposed three-turn search can effectively handle diverse fine-grained visual reasoning tasks.

\newpage
\begingroup
\ifdefined\nolinenumbers\nolinenumbers\fi
\noindent\begin{minipage}{\linewidth}
\centering
\begin{tcolorbox}[colback=promptbg, colframe=promptframe, boxrule=0.6pt, arc=2pt, left=5pt, right=5pt, top=4pt, bottom=4pt, title={\textbf{\footnotesize System Prompt for the Text-Prompting Baseline (Without $\mathbf{S}$)}}]
\scriptsize\frenchspacing
\texttt{You are a visual search assistant. Answer the user's question about the image in exactly three turns.}\\[2pt]
\texttt{Turn 1 (Propose): Look at the whole image. First write "Rationale:" followed by a short analysis of the scene and instructions for what to check in each region. Then enclose an executable Python script on I\_0 in <code>...</code> that crops and zooms all candidate regions (e.g., via normalized [0, 1000] boxes, point-centered crops, or a uniform grid for dense scenes) and stores the patches in crops.}\\[2pt]
\texttt{Turn 2 (Inspect): You will be shown one crop at a time as "Crop p:". Following your own rationale, reply with "Finding p:" and a concise description of what you see in that crop only.}\\[2pt]
\texttt{Turn 3 (Aggregate): When you see "Final answer:", combine all findings, discard distractors, resolve overlaps or duplicates between crops, and wrap the final answer in <answer>...</answer>.}
\end{tcolorbox}
\vspace{-1mm}
\captionsetup{hypcap=false}
\captionof{figure}{\textbf{System Prompt for the Text-Prompting Baseline} (\cref{subsec:ablation_studies}). Explicit text instructions that replace the learned prefix tokens $\mathbf{S}$ in Turn~1 when evaluating the frozen VLM via text prompting alone.}
\label{fig:supp_baseline_prompt}
\end{minipage}
\endgroup

\clearpage

\clearpage
\twocolumn[{%
\centering
\begin{tcolorbox}[colback=white, colframe=promptframe, boxrule=0.7pt, arc=2pt, left=5pt, right=5pt, top=5pt, bottom=5pt, title={\textbf{\footnotesize Representative Failure Cases of Our Three-Turn Search}}]
\footnotesize
\begin{minipage}[t]{0.325\linewidth}
\includegraphics[width=\linewidth]{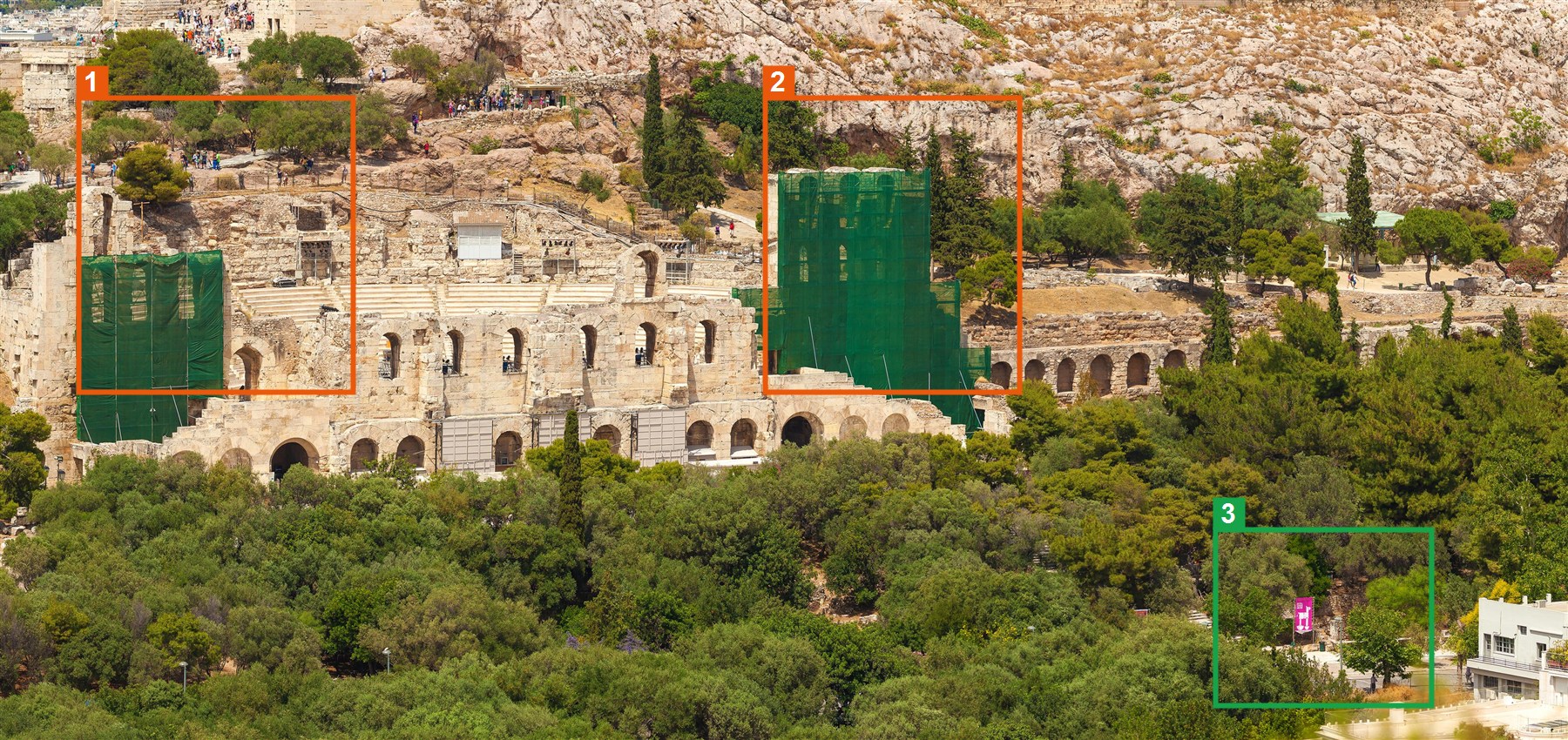}\\[2pt]
\scriptsize\raggedright \textbf{(a) Turn~1 proposal miss.} \emph{Query: What is the relative position of the pink banner compared to the green trash bin?} (HR-Bench 8K; part of the image shown). Both objects cover only a few dozen pixels of the $7680 \times 4388$ image and vanish in the global view. Turn~1 mistakes the large green scaffolding nets (orange, 1--2) for candidates and misses the actual banner and bin (green, 3), so Turns~2--3 cannot recover the answer.
\end{minipage}\hfill
\begin{minipage}[t]{0.325\linewidth}
\includegraphics[width=\linewidth, trim=0 80 0 80, clip]{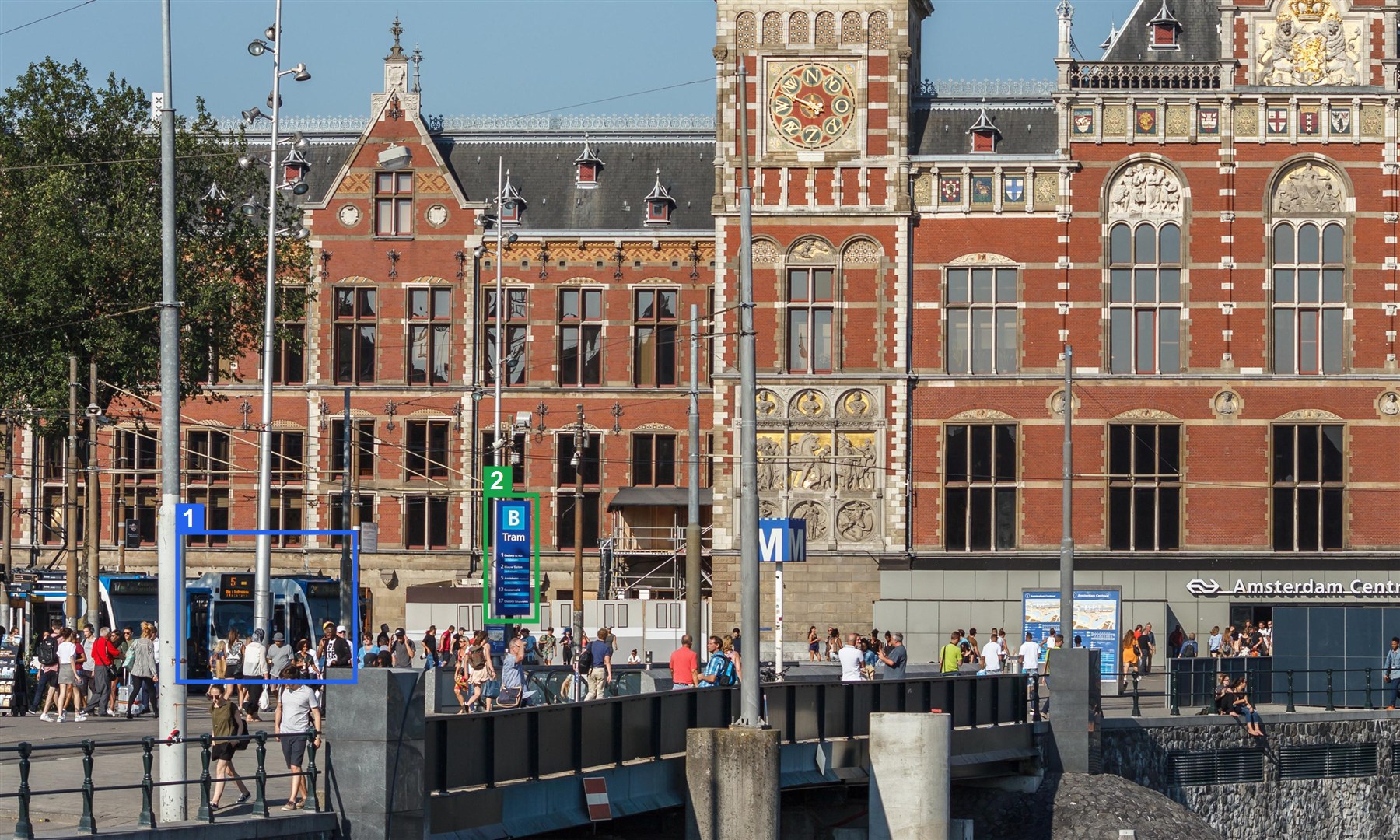}\\[2pt]
\scriptsize\raggedright \textbf{(b) Two-step lookup without a global cue.} \emph{Query: Which destination does the stop board list for the route of the tram on the left?} (part of an 8K image shown). Turn~1 correctly crops the tram (blue, 1), but its route number is only known after Turn~2. The stop board that lists this route (green, 2) is one of many small signs with no cue in the global view, so it is not proposed in Turn~1.
\end{minipage}\hfill
\begin{minipage}[t]{0.325\linewidth}
\includegraphics[width=\linewidth]{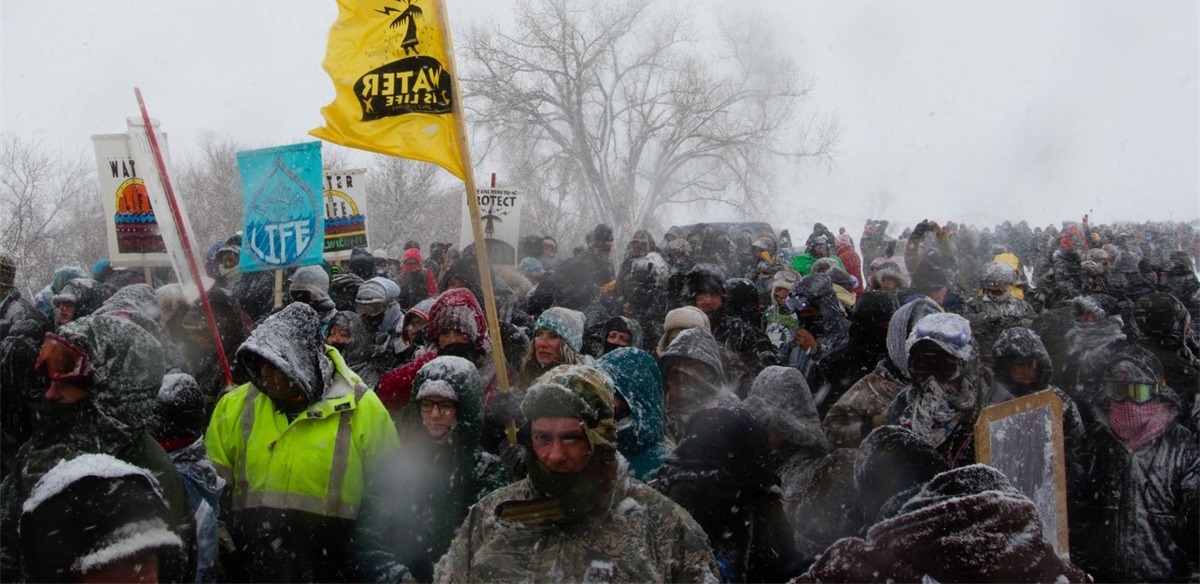}\\[2pt]
\scriptsize\raggedright \textbf{(c) Severe occlusion in Turn~2.} \emph{Query: How many people are there in the image?} (ground truth: 109). People in the dense, snow-covered crowd overlap heavily and are partly hidden by hoods and flags. Even in zoomed tiles, individual heads cannot be separated, so the tile counts are too low.
\end{minipage}
\end{tcolorbox}
\vspace{-1mm}
\captionsetup{hypcap=false}
\captionof{figure}{\textbf{Representative Failure Cases.} Challenging edge cases where errors arise from (a) sub-pixel targets in an 8K image that vanish in $\mathbf{V}_0$ alongside strong color distractors, (b) two-step lookups where the second sign has no cue in the global view until the first crop is read, and (c) severe crowd occlusion where instances overlap heavily even after zooming in.}
\label{fig:supp_failure_cases}
\vspace{1.8ex}
}]

\section{Failure Analysis, Limitations, and Future Work}
\label{supp:sec_failures}

\cref{fig:supp_failure_cases} examines three challenging edge cases that illustrate the boundary conditions of our approach. In \cref{fig:supp_failure_cases}a, the queried objects occupy only a few dozen pixels in an 8K image and disappear in $\mathbf{V}_0$, while large green scaffolding nets act as strong color distractors; Turn~1 crops those distractors and misses the true target area, leaving Turns~2--3 without the needed evidence. Note that when a target leaves no visual trace in $\mathbf{V}_0$, sequential search agents starting from the same global view face the same bottleneck short of exhaustively scanning the image. In \cref{fig:supp_failure_cases}b, whereas most relational queries expose candidate neighbors in the global layout (\cref{fig:supp_traj_multiclue}), this query requires reading a tram's route number in Crop~1 before knowing which of many identical-looking stop signs to inspect next. Finally, in \cref{fig:supp_failure_cases}c, heavy physical occlusion in a dense crowd makes overlapping people hard to separate even in zoomed tiles, reflecting the visual limit of the frozen base encoder rather than the search structure.

\vspace{0.5ex}\noindent
{\bf Limitations and Future Work.}
Across all five benchmarks, three turns already suffice to outperform multi-turn RL methods, indicating that strictly two-stage lookups without global cues (\cref{fig:supp_failure_cases}b) are rare in practice. For such edge cases, addressing them still avoids open-ended one-crop-per-turn RL: a simple extension is to let Turn~3 request one extra parallel \emph{Propose--Inspect} round ($3 \to 5$ turns) only when Turn~2 reveals a new region to check, keeping three turns for standard queries. Likewise, since Turn~2 inherits the perception of the frozen VLM (\cref{fig:supp_failure_cases}c), pairing our prefix training with stronger vision backbones directly lifts accuracy on heavily occluded scenes while preserving general capabilities. A further direction is extending our three-turn search to high-resolution video, where Turn~1 selects candidate keyframes and spatial regions from a coarse overview and Turn~2 inspects them in parallel.

\vspace{0.5ex}\noindent
{\bf Broader Impact.}
Recent work on agentic VLMs largely assumes that teaching a model multi-turn visual behaviors requires updating all model weights through large-scale SFT or reinforcement learning, which demands heavy compute and large replay mixtures to avoid forgetting general capabilities (\cref{tab:supp_baseline_data}). Our results point to a different view: modern pre-trained VLMs already possess the core visual, coding, and reasoning abilities needed for complex tasks, and mainly lack a way to coordinate them across turns. In this light, prefix tuning acts as a lightweight ``memory implant'' over full-model fine-tuning, injecting a tiny procedural memory ($20.5$--$28.7\text{ KB}$) that guides a frozen VLM through a structured workflow while leaving its weights and general abilities untouched. In practice, a single frozen VLM deployment can swap compact skill prefixes on demand without storing separate multi-gigabyte checkpoints, while keeping intermediate steps ($r$, $\mathcal{C}$, and $\{\mathbf{a}^{(p)}\}_{p=1}^P$) explicit and easy to verify. Beyond visual search, we believe this approach can be extended to a broad range of visual tasks for VLMs, such as multi-page document reading, GUI navigation, and medical or satellite image analysis.

\end{document}